\documentclass[sigconf]{acmart}
\usepackage{afterpage}
\usepackage{xspace}
\usepackage{subcaption}
\usepackage{array}
\usepackage{bbding} 
\usepackage{graphicx}
\usepackage{multirow}
\usepackage{makecell}
\usepackage{colortbl}
\newcommand{\method}{\texttt{MIGE}\xspace}
\AtBeginDocument{%
  }

\copyrightyear{2025}
\acmYear{2025}
\setcopyright{acmlicensed}
\acmConference[MM '25]{Proceedings of the 33rd ACM International Conference on Multimedia}{October 27--31, 2025}{Dublin, Ireland}
\acmBooktitle{Proceedings of the 33rd ACM International Conference on Multimedia (MM '25), October 27--31, 2025, Dublin, Ireland}\acmDOI{10.1145/3746027.3755811}
\acmISBN{979-8-4007-2035-2/2025/10}
\acmSubmissionID{6167}

\begin{document}

\title[MIGE]{MIGE: Mutually Enhanced Multimodal Instruction-Based \\ Image Generation and Editing}

% CCS Concepts



%Keywords
% Keywords. The author(s) should pick words that accurately describe
% the work being presented. Separate the keywords with commas.


% A "teaser" image appears between the author and affiliation
% information and the body of the document, and typically spans the
% page.



\begin{teaserfigure}
\centering
  \includegraphics[width=0.92\textwidth,height = 7.1 cm]{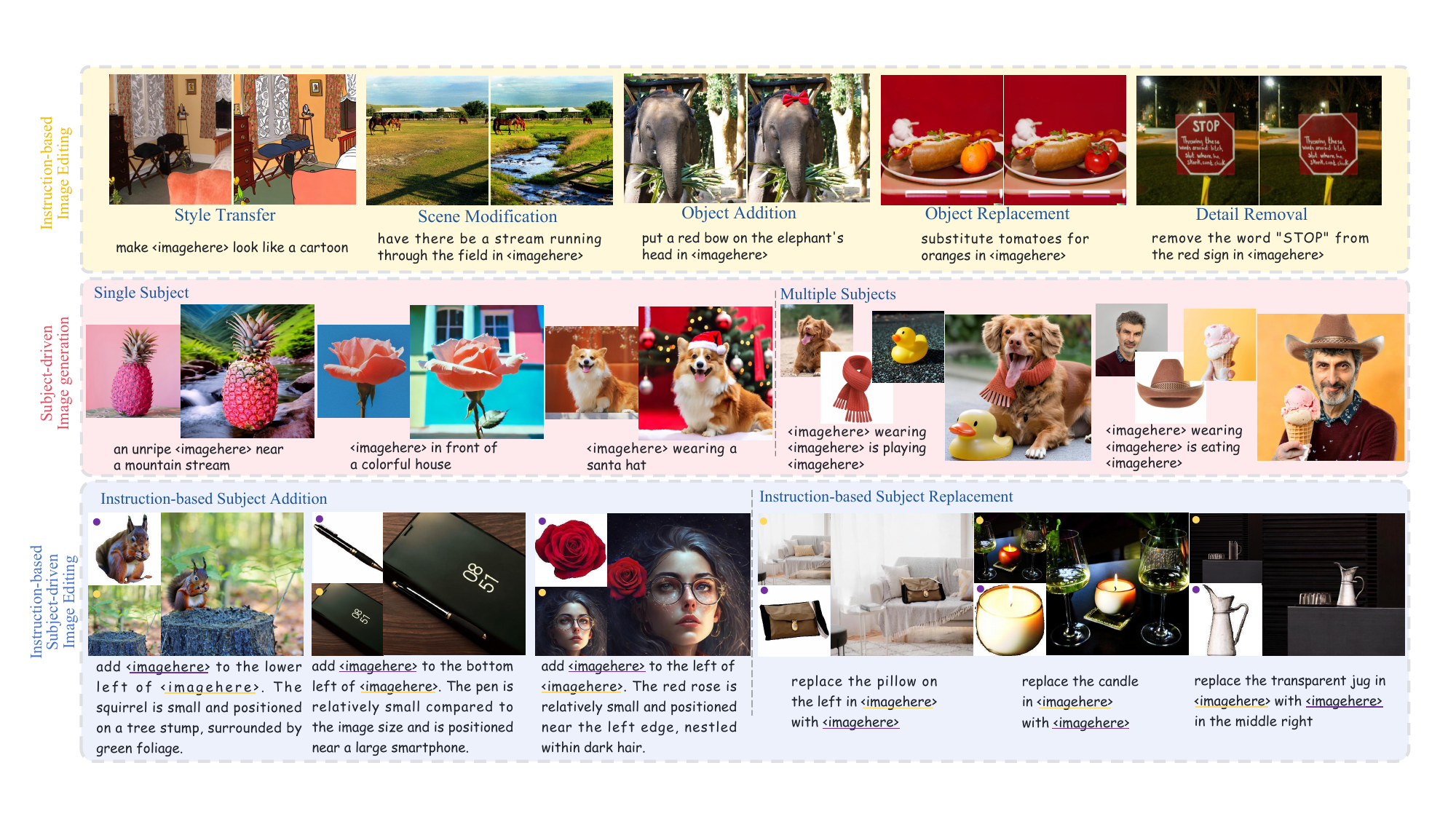}
  \caption{\textbf{Demonstrating the comprehensive capabilities of MIGE}, a unified framework that excels in subject-driven generation and instruction-based editing, while achieving the best performance on instruction-based subject-driven editing.}
  \label{fig:show}
\end{teaserfigure}
% \affiliation{%
\author{Xueyun Tian}
\email{tianxueyun23z@ict.ac.cn}
\affiliation{%
  \institution{State Key Laboratory of AI Safety, Institute of Computing Technology, Chinese Academy of Sciences}
  \city{Beijing}
  \country{China}
}
\affiliation{%
  \institution{University of Chinese Academy of Sciences}
  \city{Beijing}
  \country{China}
}

\author{Wei Li}
\email{weili.ucas.ict@gmail.com}
\affiliation{%
  \institution{State Key Laboratory of AI Safety, Institute of Computing Technology, Chinese Academy of Sciences}
  \city{Beijing}
  \country{China}
}

\author{Bingbing Xu}
\email{xubingbing@ict.ac.cn}
\affiliation{%
  \institution{State Key Laboratory of AI Safety, Institute of Computing Technology, Chinese Academy of Sciences}
  \city{Beijing}
  \country{China}
}

\author{Yige Yuan}
\email{yuanyige20z@ict.ac.cn}
\affiliation{%
  \institution{State Key Laboratory of AI Safety, Institute of Computing Technology, Chinese Academy of Sciences}
  \city{Beijing}
  \country{China}
}
\affiliation{%
  \institution{University of Chinese Academy of Sciences}
  \city{Beijing}
  \country{China}
}

\author{Yuanzhuo Wang}
\email{wangyuanzhuo@ict.ac.cn}
\affiliation{%
  \institution{State Key Laboratory of AI Safety, Institute of Computing Technology, Chinese Academy of Sciences}
  \city{Beijing}
  \country{China}
}

\author{Huawei Shen}
\email{shenhuawei@ict.ac.cn}
\affiliation{%
  \institution{State Key Laboratory of AI Safety, Institute of Computing Technology, Chinese Academy of Sciences}
  \city{Beijing}
  \country{China}
}
\affiliation{%
  \institution{University of Chinese Academy of Sciences}
  \city{Beijing}
  \country{China}
}

\begin{abstract}
Despite significant progress in diffusion-based image generation, subject-driven generation and instruction-based editing remain challenging. 
% 两个任务分开做的缺点
Existing methods typically treat them separately, struggling with limited high-quality data and poor generalization.
% 两个任务的相似性
However, both tasks require capturing complex visual variations while maintaining consistency between inputs and outputs.
%提framework
Inspired by this,
% Therefore, 
we propose \method, a unified framework that standardizes task representations using multimodal instructions. It first treats subject-driven generation as creation on a blank canvas and instruction-based editing as modification of an existing image, establishing a shared input-output formulation, then introduces a novel multimodal encoder that maps free-form multimodal instructions into a unified vision-language space, integrating visual and semantic features through a feature fusion mechanism.
%好处
This unification enables joint training of both tasks, providing two key advantages:
(1) \textbf{Cross-Task Enhancement}: by leveraging shared visual and semantic representations, joint training improves instruction adherence and visual consistency in both subject-driven generation and instruction-based editing.
(2) \textbf{Generalization}: learning in a unified format facilitates cross-task knowledge transfer, enabling \method to generalize to novel compositional tasks, including instruction-based subject-driven editing.
Experiments show that \method excels in both subject-driven generation and instruction-based editing while setting a SOTA in the new task of instruction-based subject-driven editing. 
Code and model have been publicly available at \href{https://github.com/Eureka-Maggie/MIGE}{this link}.
\end{abstract}

\maketitle

\section{Introduction}
% Paragraph 1: 两个任务
% Recent advances in diffusion models \cite{rombach2022high,peebles2023scalable,flux2023,esser2024scaling} have greatly advanced customized image generation, enabling a broad range of applications in subject-driven image generation \cite{ruiz2023dreambooth,ye2023ip,pan2023kosmos} and instruction-based editing \cite{brooks2023instructpix2pix,sheynin2024emu,fu2023guiding,han2024ace}. The former focuses on preserving a given subject, and the latter emphasizes flexible modifications based on textual instructions. They have attracted increasing attention due to their ability to enable personalization.

Recent advances in diffusion models\cite{rombach2022high,peebles2023scalable,flux2023,esser2024scaling} have significantly advanced the development of customized image generation, enabling its widespread adoption in both subject-driven generation\cite{ruiz2023dreambooth,ye2023ip,li2024blip,pan2023kosmos,kang2025flux,kim2024learning,wu2025core,liang2025idea,peng2024dreambench++} and instruction-based editing\cite{brooks2023instructpix2pix,sheynin2024emu,fu2023guiding,han2024ace}. The former focuses on preserving a given subject, while the latter emphasizes flexible modifications guided by textual instructions. Their capability to generate personalized and controllable results demonstrates substantial practical value, thereby attracting growing research attention.
 
% Paragraph 2: 分开做缺点,共同指向了 单一建模数据缺乏且泛化性差 
Existing approaches \cite{li2024blip,pan2023kosmos,hu2024instruct,huang2024smartedit} typically treat these two tasks independently and address them separately by tailoring specific input-output formats to each. However, this results in limited generalization capability and necessitates extensive training data.
Specifically, subject-driven generation aims to preserve subject identity while adhering to textual instructions. Fine-tuning methods, like DreamBooth \cite{ruiz2023dreambooth}, require separate training for each subject during testing, which limits generalization. In contrast, zero-shot methods like KOSMOS-G \cite{pan2023kosmos} rely on multimodal alignment but struggle with instruction following, mainly due to the scarcity of high-quality multimodal training data, especially for multi-subject inputs. Instruction-based editing, exemplified by InstructPix2Pix \cite{brooks2023instructpix2pix}, modifies images based on instructions. These methods require large, high-quality editing datasets to handle diverse instructions but struggle with maintaining local-to-global consistency.

The above separate processing methods can only rely on limited data and single-task optimization objectives, which hinders their ability to adapt to diverse subjects or editing requirements, and lack cross-task consistency, preventing a unified understanding of the tasks. To address these challenges, we shift our perspective from separate processing to the objectives of these two tasks. These two tasks share a fundamental principle: maintaining visual consistency between inputs and outputs while capturing variations guided by instructions. They also follow a similar input-output paradigm with complementarities: subject-driven generation emphasizes subject accuracy, while instruction-based editing manipulates images without altering irrelevant identities. These limitations and complementarities highlight the need for a unified framework that takes advantage of the strengths of both tasks to improve performance.

%Both tasks face significant challenges due to limited data, which hinders their ability to adapt to diverse subjects or editing requirements. Furthermore, they are typically trained independently, lacking cross-task consistency, which prevents a unified understanding of the tasks. Despite these challenges, the two tasks share a fundamental principle: maintaining visual consistency between inputs and outputs while capturing variations guided by instructions. They also follow a similar input-output paradigm with complementary data focuses—subject-driven generation emphasizes subject accuracy, while instruction-based editing manipulates images without altering irrelevant identities. These limitations and complementarities highlight the need for a unified framework that leverages the strengths of both tasks for improved performance.

% Paragraph 4: The Proposed Unified Framework
Inspired by the above analysis,
%Therefore, 
we propose a unified framework, \method, which leverages \textbf{M}ultimodal \textbf{I}nstructions to unify subject-driven image \textbf{G}eneration and instruction-based \textbf{E}diting. The vision-language representation enables substituting both entities and entire images in text prompts with their visual counterparts, allowing flexible task and instruction combinations.
Structurally, we model subject-driven generation as creating an image on a blank canvas and instruction-based editing as modifying an existing image. This coherent input-output mapping simplifies the process while enabling both tasks to reinforce each other through joint training.
Moreover, this integrated approach fosters emergent compositional capabilities beyond what either task can achieve alone.

% Paragraph 5: Encoder Improvements, and Data Construction
Building on our unified framework, we address two critical challenges. First, for multimodal instruction encoding, existing methods \cite{li2023blip, pan2023kosmos, li-etal-2024-unimo} mainly rely on CLIP vision encoders \cite{radford2021learning} to extract semantic features, which is insufficient for preserving fine-grained subject details. To overcome this limitation, we propose a multimodal encoder with a feature fusion mechanism that integrates VAE \cite{kingma2013auto} visual features into semantic tokens, effectively capturing both detailed visual information and semantics.
Secondly, we enhance compositionality through joint training, facilitating instruction-based subject-driven editing. To further optimize performance in this complex task, we develop a novel data construction pipeline based on a Multimodal Large Language Model (MLLM). This pipeline autonomously generates diverse multimodal instructions and corresponding output images. Moreover, to address the absence of an evaluation benchmark for this new composition task, we introduce MIGEBench. This specialized benchmark evaluates compositionality in terms of subject preservation and instruction adherence, providing a comprehensive assessment of the new task.

% Paragraph 6: Experimental Results and Case Studies

\begin{figure}[t]
    \centering
    \includegraphics[width=0.35\textwidth,height=5.2cm]{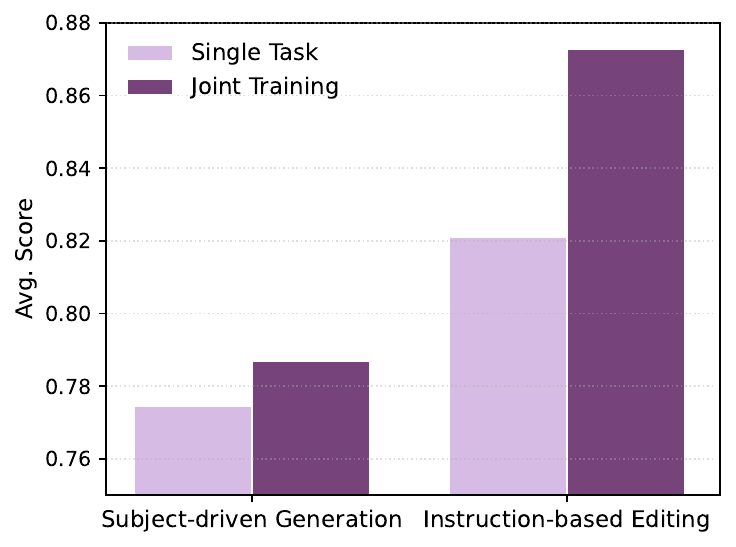} 
    % \caption{Demonstration of joint training effectiveness. We compute the average DINO and CLIP-I scores for subject-driven generation and instruction-based editing under single-task and joint training settings. Joint training significantly improves input-output similarity across both tasks.}
     \caption{Joint training significantly improves input-output similarity across both tasks. We show the average DINO and CLIP-I scores for subject-driven generation and instruction-based editing under single-task and joint training.}
    \label{fig:bar}
\end{figure}

Our experiments demonstrate that joint training under the \method framework enables mutual enhancement between subject-driven generation and instruction-based editing, leading to significant performance gains over training on individual tasks. As shown in Figure \ref{fig:bar}, joint training markedly improves input-output consistency across both tasks, especially in editing, with scores rising from 0.821 to 0.873.
Moreover, \method achieves state-of-the-art results on MIGEBench, showcasing its fine-grained controllability in emerging compositional tasks. Extensive case studies (Figures~\ref{fig:compare_sub}--\ref{fig:compare_new}) further demonstrate the ability of \method to generate precise and consistent outputs, highlighting its effectiveness across diverse scenarios.

% Our experiments demonstrate that joint training under the \method framework enables mutual reinforcement between subject-driven generation and instruction-based editing by leveraging complementary data, leading to significant performance gains over training on individual tasks. As shown in Figure \ref{fig:bar}, joint training significantly enhances input-output consistency across both tasks, especially in editing scenario, with scores rising from 0.821 to 0.873. %Specifically, we observe improved instruction adherence on DreamBench \cite{ruiz2023dreambooth} and enhanced detail preservation on EmuEdit and MagicBrush \cite{zhang2024magicbrush} test sets. 
% Furthermore, \method achieves state-of-the-art results on MIGEBench, showcasing its fine-grained controllability in the emerging compositional tasks. Extensive case studies (Figures \ref{fig:compare_sub}, \ref{fig:compare_edit}, and \ref{fig:compare_new}) further demonstrate the ability of \method to generate precise and consistent outputs, highlighting its effectiveness across diverse scenarios.

In summary, our contributions include: 
\begin{itemize}

\item \textbf{Promising perspective of task mutual enhancement}: We shift our perspective from traditional separate processing to mutual enhancement. Then, we design a unified input-output format for joint training on subject-driven generation and instruction-based editing, validating that joint training enables task enhancement.

\item \textbf{Novel framework for Compositional Capability}: We propose \method, a unified framework integrating two mainstream tasks in controllable image generation. Joint training further enables the compositional capability of instruction-based subject-driven editing. We also introduce a novel data pipeline and MIGEBench for comprehensive evaluation.

% \item \textbf{First validation of task mutual enhancement}: We design a unified input-output format for joint training on subject-driven generation and instruction-based editing, and validate that joint training enables task enhancement.

% \item \textbf{Compositional Capability}: We propose \method, a unified framework integrating two mainstream tasks in controllable image generation. Joint training further enables the compositional capability of instruction-based subject-driven editing. We also introduce a novel data pipeline and the MIGEBench benchmark for comprehensive evaluation.

\item \textbf{Strong Performance with Extensive Experiments}: Extensive results show that \method performs competitively on subject-driven generation and instruction-based editing, with scores rising from 0.821 to 0.873, and sets a new state-of-the-art in instruction-based subject-driven editing.
\end{itemize}

\section{Related Work}
In this section, we discuss the most relevant studies. A more comprehensive discussion can be found in the Supplementary Material. %\ref{sec:related work}.

\paragraph{Universal generative model}
Unifying tasks in image generation remains challenging. Some methods \cite{li2025visualcloze,huang2024group} focus on bridging image understanding and generation, such as SEED-X \cite{ge2024seed}, Emu2 \cite{sun2024generative}, and WeGen \cite{huang2025wegen}, which use autoregressive models with auxiliary diffusion modules. UnifiedMLLM \cite{li2024unifiedmllm} employs MLLMs to learn shared representations and route to task-specific models. However, these approaches either emphasize understanding over generation or rely on complex pipelines.
Others aim to unify generation and editing. Pixwizard \cite{lin2024pixwizard} uses text-only instructions and task vectors. Instruct-Imagen \cite{hu2024instruct} supports multimodal inputs but performs poorly on editing. ACE \cite{han2024ace} is limited to editing via transformer-based conditioning. OmniGen \cite{xiao2024omnigen} cannot distinguish reference images. OmniControl \cite{tan2024ominicontrol} merges all conditions with noise, while UniReal \cite{chen2024unireal} adds finer control but requires complex prompts. RealGeneral \cite{lin2025realgeneral} enhances attention, and OnePic \cite{tao2024we} depends on multiple references and masks.
Overall, prior work often sacrifices flexibility or generality, combining tasks without leveraging their intrinsic connections. In contrast, we propose \method, which jointly trains on subject-driven generation and instruction-based editing, revealing their mutual benefits.

\paragraph{Subject-Driven Image Editing}
Subject-driven image editing enables modifications based on a user-specified subject, typically including addition or replacement. Existing methods often rely on multiple reference images or precise cues. DreamEdit \cite{li2023dreamedit} and DreamMix \cite{yang2024dreammix} require multiple images for fine-tuning and iterative inpainting. PBE \cite{yang2023paint} uses compressed reference embeddings, while TIGIC \cite{li2025tuning} embeds subjects into predefined backgrounds. MADD \cite{he2024affordance} supports diverse controls via masks or coordinates. AnyDoor \cite{chen2024anydoor}, AnyEdit \cite{yu2024anyedit}, MimicBrush \cite{chen2024zero} and BlobCtrl \cite{li2025blobctrl} require masks or additional range blobs. FreeEdit \cite{he2024freeedit} is limited to instruction templates and class-aware features. Overall, these approaches depend on detailed inputs and lack flexibility in handling diverse instructions for controllable generation.

\begin{figure}[t]
    \centering
    \includegraphics[width=0.45\textwidth,height=5cm]{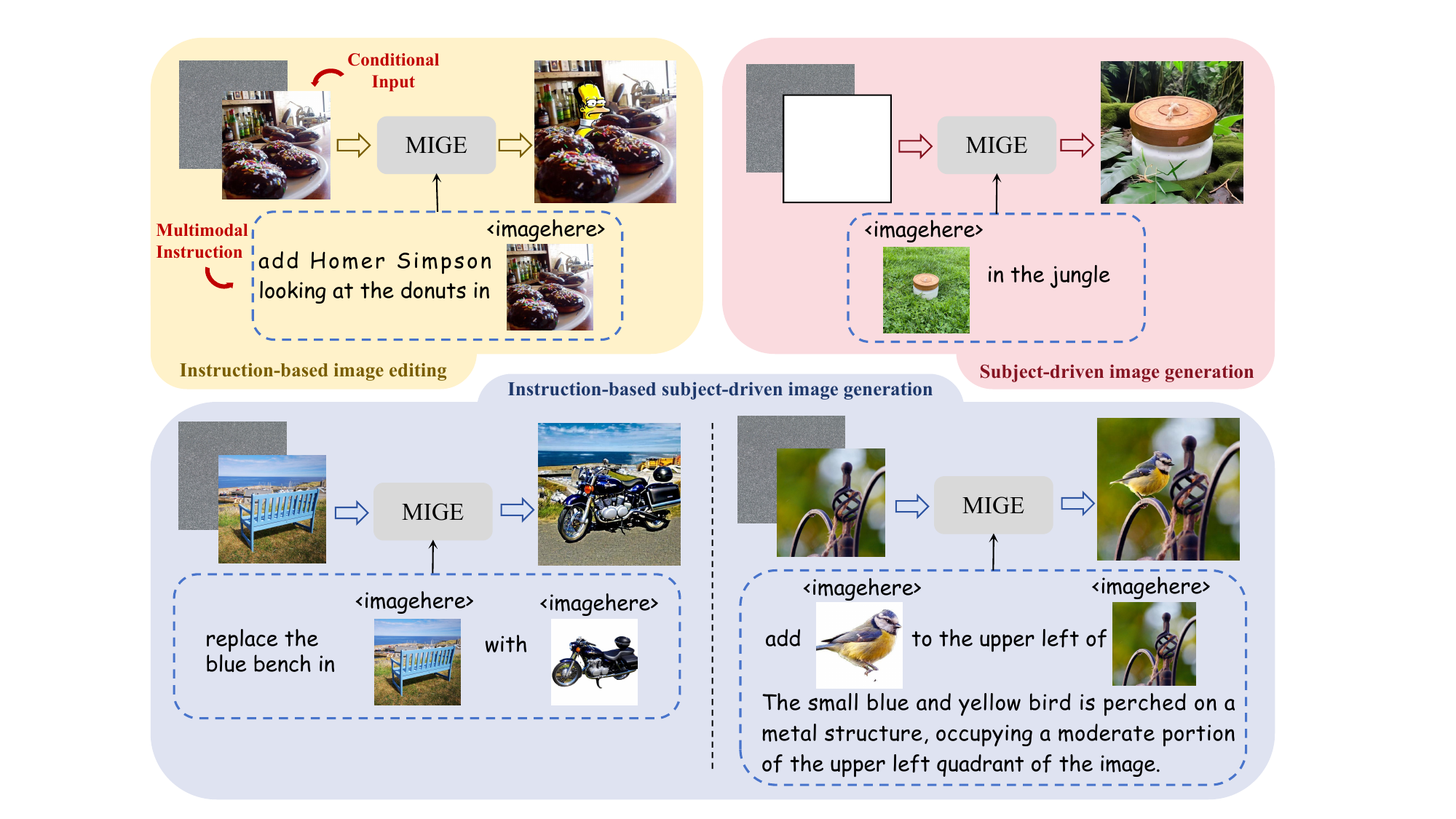} 
    \caption{Demonstration of MIGE as a unified framework for processing multimodal instructions and conditional inputs across diverse scenarios.}
    \label{fig:format}
\end{figure}
\begin{figure*}[ht]
    \centering
    \includegraphics[width=0.8\textwidth,height=6.2cm]{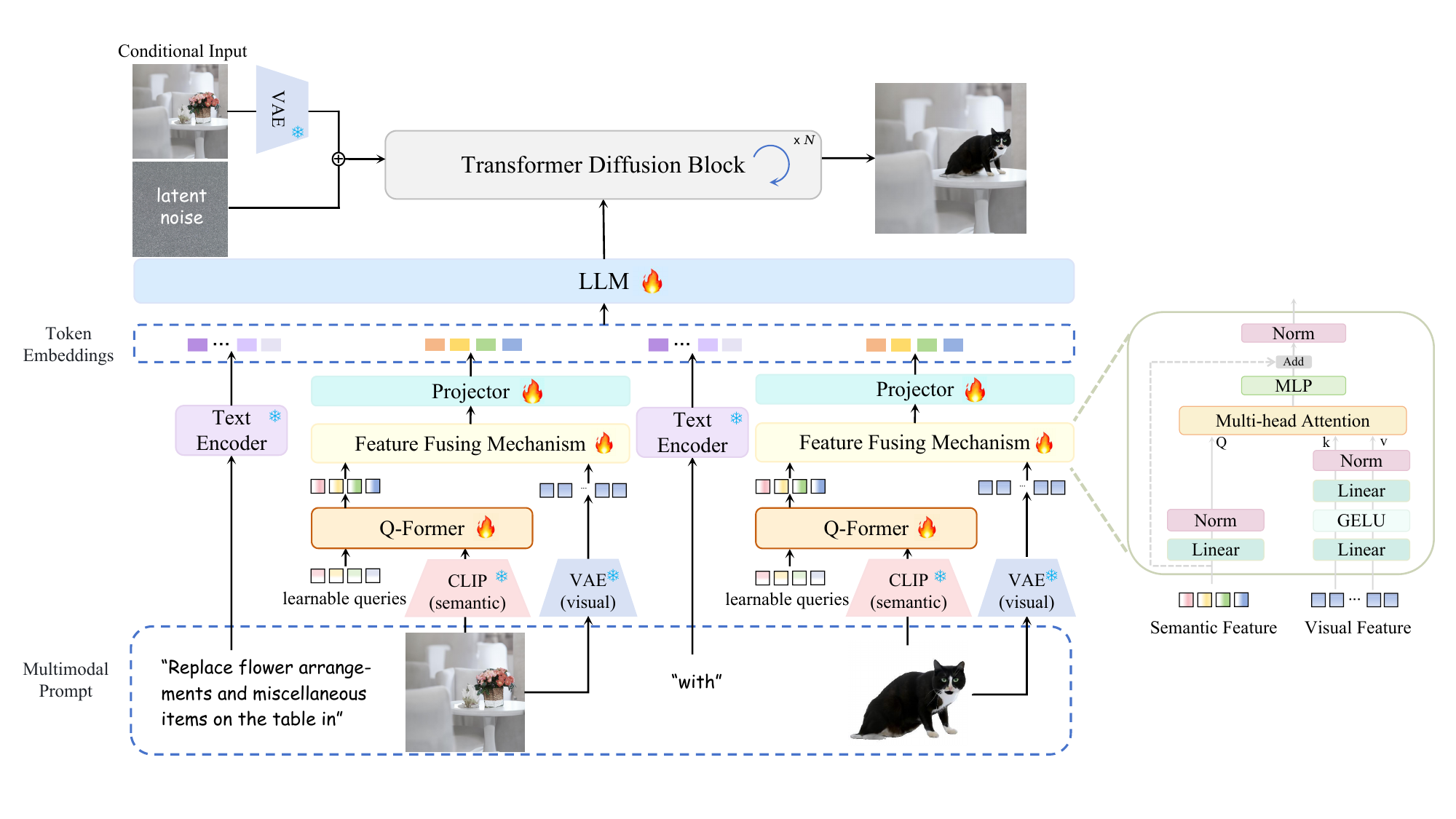} 
    \caption{\textbf{Overall framework of MIGE.} MIGE consists of two components: a multimodal encoder for processing multimodal instructions and a transformer-based diffusion model for modeling input-output relationships. The encoder incorporates a feature fusion mechanism to integrate visual and semantic features from reference image.}
    \label{fig:model_new}
\end{figure*}

\section{Method}
%The architecture of \method, as shown in Figure \ref{fig:model_new}, consists of two main components: a MLLM for encoding multimodal instructions and a diffusion model for image generation under multimodal conditions. The MLLM encodes interleaved image-text inputs into a unified semantic space, comprising a VAE, ViT, Q-Former, linear projection layer, and a T5 model. Each image is represented by 32 tokens, which are projected into the semantic space and encoded alongside text tokens to form a unified multimodal condition. The diffusion model then utilizes this condition, along with the conditional input, to generate images accordingly.
In this section, we first shift our view from separate training to mutual enhancement. Guided by this new perspective, we first introduce the unified framework \method, followed by details of the architecture design and joint training approach.
\subsection{Unified framework MIGE}
Current methods treat subject-driven generation and instruction-based image editing separately, hindering performance due to data limitations and poor generalization. Since both aim to preserve visual consistency while incorporating instructed changes, unifying them allows mutual improvement. Joint training on diverse data enhances subject preservation and instruction following beyond the capabilities of individual task-specific models.

Therefore, we propose \method to unify the two tasks for joint training. By using multimodal instructions as a unified task representation, it supports flexible combinations and provides multimodal guidance. Additionally, we use conditional input to structurally unify the tasks, enhancing visual consistency. Their combination not only provides rich visual and instructional information but also naturally represents different tasks. We demonstrate the input-output format of \method in Figure \ref{fig:format}.

\subsubsection{Unified Multimodal Instruction}
To enable joint training across multiple tasks, a unified task representation is essential. We introduce multimodal instructions composed of interleaved images and text, which provide both visual references and textual guidance for various controllable generation tasks. As shown in Figure \ref{fig:format}, ``<imagehere>'' serves as a placeholder, sequentially replaced by input images, which can be the reference subject or the entire scene, complementing the semantics to form interleaved expressions. This unified approach effectively accommodates subject-driven generation and instruction-based editing while remaining extensible to the more complex combination task. 

\subsubsection{Unified Conditional Input}
We design a unified conditional input format to distinguish tasks while enabling shared capabilities. As shown in Figure~\ref{fig:format}, conditional inputs vary across tasks. For instruction-based editing, the VAE-encoded source image is used to preserve fine details of the original image, inform the model to edit only the target region, as done in prior editing methods \cite{brooks2023instructpix2pix,zhang2024magicbrush}. To standardize the input format, we concatenate an all-zero tensor for subject-driven generation to indicate generation from scratch with visual constraints from the multimodal instruction. This design unifies the input format, ensures input-output consistency, and facilitates extensibility to new tasks within a unified framework.

%This design ensures input-output consistency in editing and facilitates extensibility to new tasks within a unified framework.
%By concatenating different conditions, we differentiate tasks and capture task-specific nuances.

\subsection{Architecture}
The architecture of \method, as shown in Figure \ref{fig:model_new}, consists of two main components: a multimodal encoder for processing multimodal instructions and a transformer-based diffusion model \cite{peebles2023scalable} for modeling input-output relationships. The diffusion model takes concatenated latent noise and conditional input along the channel dimension as input, performing controllable generation under the control of multimodal conditions. To further enhance the integration of visual and semantic information from the reference image, we introduce a novel feature fusion mechanism in the encoder.

\subsubsection{Multimodal Encoder}
To map the multimodal instructions into a unified vision-language semantic space, we design a multimodal encoder comprising a Large Language Model (LLM) and an image feature encoding component, which includes a pretrained VAE encoder \cite{kingma2013auto} for visual feature extraction, a pretrained ViT from EVA-CLIP \cite{fang2023eva} for semantic feature extraction, a Q-Former \cite{li2023blip}, and a linear projection layer. Each image is represented by 32 tokens, which are then encoded by the LLM alongside text tokens to form a unified multimodal condition.

Prior works \cite{li2024blip,pan2023kosmos,li-etal-2024-unimo} focus on extracting semantics from reference images but lack the fine-grained details needed to preserve subject-specific features.
To address this, we propose a feature fusion mechanism that combines the strengths of different visual encoders. ViT serves as a semantic feature extractor, while a VAE encoder acts as a visual feature extractor, leveraging its ability to compress and reconstruct images. As shown in Figure \ref{fig:model_new}, we use the ViT semantic features compressed by the Q-Former, denoted as $f_s$, as guidance to adaptively incorporate the visual features $f_v$ extracted by the VAE. Inspired by TokenPacker~\cite{li2024tokenpacker}, we represent rich features with fewer tokens, and apply ReLU~\cite{glorot2011deep} to enhance stability. The fusion mechanism is expressed as: $
  f_{\text{img}} = f_s + \text{MLP}(\text{Attn}(f_s, f_v))
$ where $\text{Attn}(f_s, f_v) = \frac{Q(f_s) \cdot K(f_v)}{\sqrt{d}} V(f_v)
$. Here $d$ denotes the output dimension of the features, \( Q \) is a linear layer, and both \( K \) and \( V \) are two-layer networks, where \( K(\cdot) = V(\cdot) = \text{Linear}(\text{GELU}(\text{Linear}(\cdot))) \).

This fusion mechanism captures both semantic and visual information without introducing extra image tokens. The multimodal encoder compresses each image into 32 fused tokens, which replace the ``<imagehere>'' placeholder in the interleaved instruction. The LLM then encodes the entire prompt into a unified vision-language representation to guide the diffusion model.

% \begin{figure*}[t]
% \centering
% \subfloat[Subject Addition Dataset Construction Pipeline]{
% \label{fig:add_pipeline}
% \includegraphics[width=0.41\textheight]{figs/add_pipeline.pdf}}
% %\hspace{20pt}
% \subfloat[Subject Replacement Dataset Construction Pipeline]{
% \label{fig:replace_pipeline}
% \includegraphics[width=0.44\textwidth]{figs/replace_pipeline.pdf}}
% \caption{Data construction pipelines for instruction-based subject-driven image editing.}
% \label{1}
% \end{figure*}
% \begin{figure*}[t]
%     \centering
%     % 第一个子图
%     \begin{subfigure}{0.45\textwidth}
%         \includegraphics[width=0.95\linewidth]{figs/add_pipeline.pdf}
%         \caption{Subject Addition Dataset Construction Pipeline}
%         \label{fig:add_pipeline}
%     \end{subfigure}
%     % 第二个子图
%     \begin{subfigure}{0.45\textwidth}
%         \includegraphics[width=0.88\linewidth]{figs/replace_pipeline.pdf}
%         \caption{Subject Replacement Dataset Construction Pipeline}
%         \label{fig:replace_pipeline}
%     \end{subfigure}

%     \caption{Data construction pipelines for instruction-based subject-driven image editing.}
%     \label{fig:data_construction}
% \end{figure*}

\begin{figure*}[t]
    \centering
    % 第一个子图
    \begin{subfigure}{0.452\textwidth}
        \includegraphics[width=0.95\linewidth]{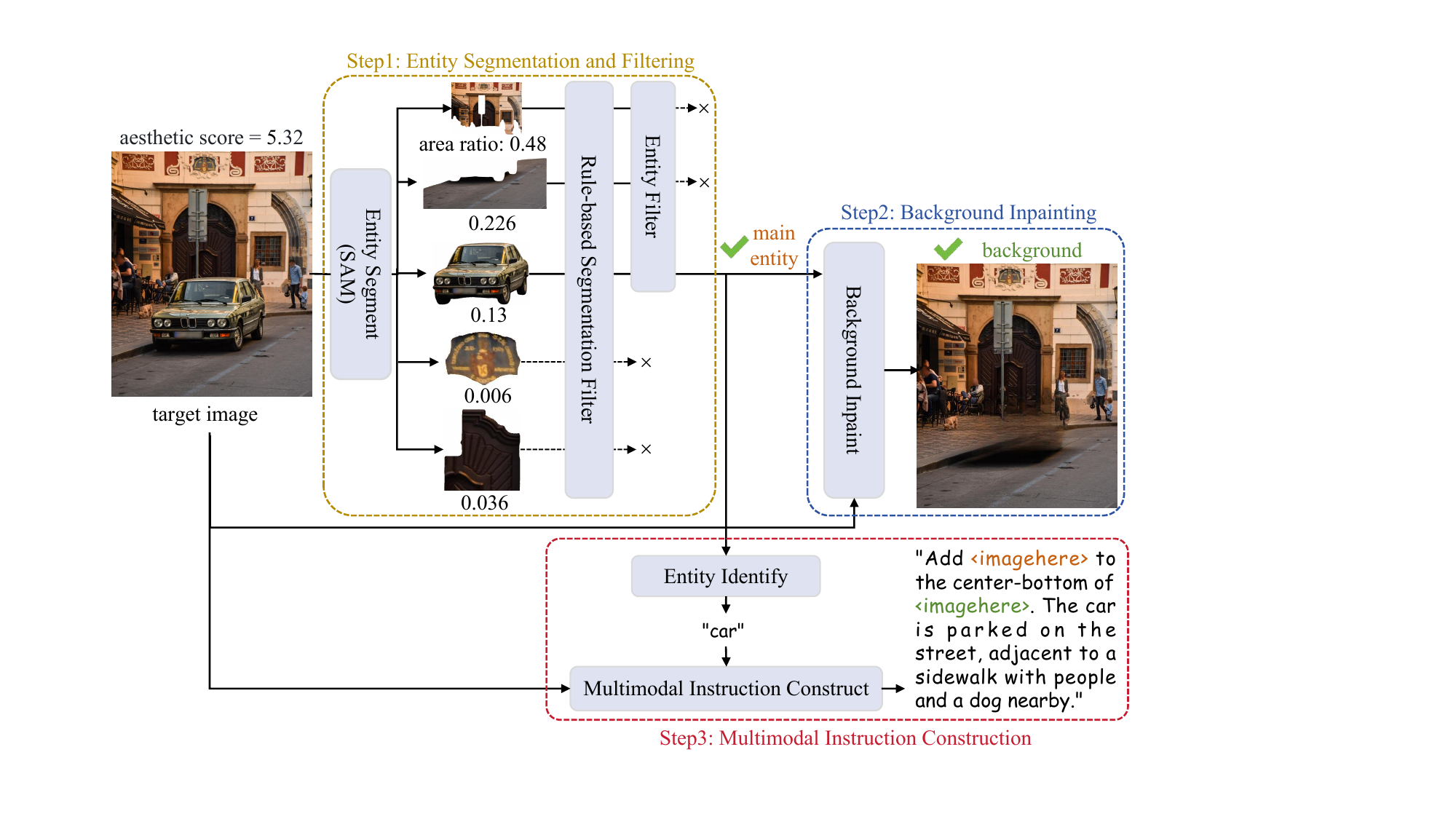}
        \caption{Subject Addition Dataset Construction Pipeline}
        \label{fig:add_pipeline}
    \end{subfigure}
    % 第二个子图
    \begin{subfigure}{0.452\textwidth}
    \centering
        \includegraphics[width=0.89\linewidth]{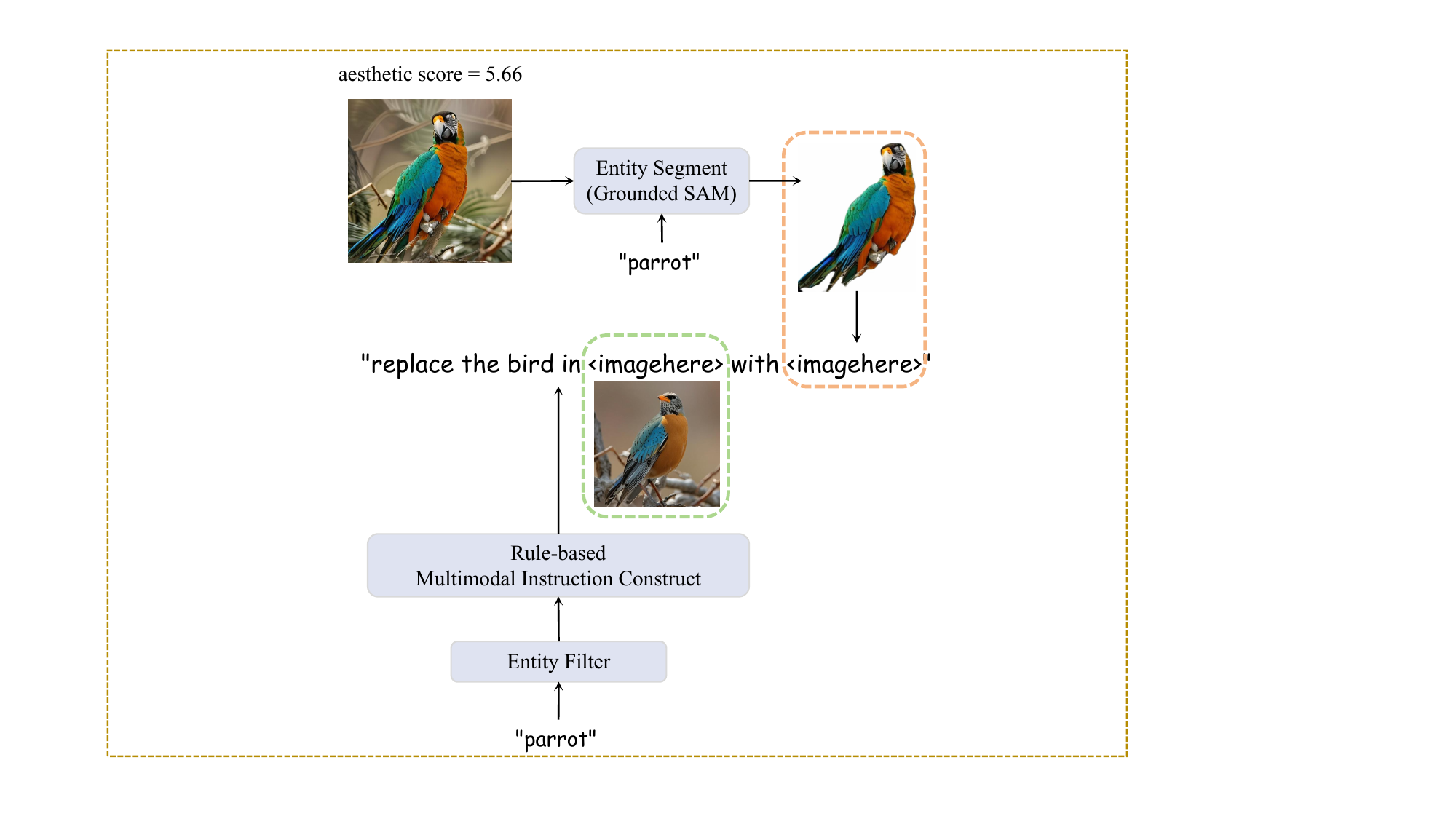}
        \caption{Subject Replacement Dataset Construction Pipeline}
        \label{fig:replace_pipeline}
    \end{subfigure}

    \caption{Data construction pipelines for instruction-based subject-driven image editing.}
    \label{fig:data_construction}
\end{figure*}

\subsection{Joint Training}
Multimodal instructions and conditional input unify task representation and input-output formats, enabling joint training. We fine-tune \method on data from all tasks to enhance cross-task synergy. Except for the two image encoders, all parameters are jointly trained to align the conditional space of the diffusion model with the multimodal encoder, as shown in Figure \ref{fig:model_new}. This approach improves task coordination and consistency across modalities.

%%%%%%%%%%%%%%%%%%%%%%%%%%%%%%%%%%%%%%%%%%%%%%%%%%%%
\subsubsection{Data Construction}
Joint training enables multi-task learning, balancing subject preservation and instruction, followed by modeling task relationships. We create a multi-task dataset for joint multimodal instruction tuning, covering subject-driven image generation, instruction-based image editing, and instruction-based subject-driven image generation.

For subject-driven image generation, we follow the data construction methods of KOSMOS-G \cite{pan2023kosmos} and UNIMO-G \cite{li-etal-2024-unimo}, using an LLM to extract entities from captions, which are then fed into Grounded SAM \cite{ren2024grounded} for segmentation. Subjects200k dataset from OmniControl \cite{tan2024ominicontrol} is also incorporated for better object preservation. For instruction-based editing, we filter existing datasets and use a rule-based strategy to construct multimodal instructions.

Instruction-based subject-driven image generation is an emerging task involving two subtasks: instruction-based subject addition and subject replacement, which allows users to add or replace a specified subject in an image using multimodal instructions. However, there is no sufficient dataset available for this task.

% addition
For instruction-based subject addition, we propose a pipeline inspired by SAM-FB \cite{he2024affordance}, as shown in Figure \ref{fig:add_pipeline}. Starting with the SA-1B \cite{kirillov2023segment} dataset, we construct the source image and a multimodal instruction for input-output pairs. We use SAM \cite{kirillov2023segment} to segment annotated entities, filter and retain the main subject using an MLLM, inpaint the remaining parts to create the background, and then combine the subject name with the target image to generate the multimodal instruction via GPT-4o \cite{hurst2024gpt}. Due to resource limitations, we process a part of the SA-1B dataset and obtain about 200k samples, but the pipeline can be scaled to generate more.
% replacement
For instruction-based subject replacement, we filter from existing editing data, use Grounded SAM for subject segmentation, and construct multimodal instructions to form input-output pairs, as shown in Figure~\ref{fig:replace_pipeline}. We also introduce virtual try-on data constructed using IDM-VTON \cite{choi2024improving}, resulting in approximately 110k samples.
More details on training data construction are in Supplementary Material. %~\ref{sec:appen_data construction}.

%Specifically, for the subject addition task,  First, we filter entities of appropriate size using predefined rules, then use an MLLM to extract the main entity. The target image and the extracted entity name are fed to GPT-4o \cite{hurst2024gpt} to construct a multimodal instruction, while the image without the chosen entity is sent for inpainting to generate the background. 

%For subject replacement, we construct both the multimodal instructions and entity extraction based on predefined rules. The target image and extracted entity names are then fed into Grounded SAM to obtain entity segmentation, 
%IDM-VTON \cite{choi2024improving} can be used to construct a portion of virtual try-on data, but we aim for the model to learn a more general generation and editing capability. To address this gap, we design MLLM-based data construction pipelines.

% \begin{figure*}[t]
%     \centering
%     % 第一个子图
%     \begin{subfigure}{0.45\textwidth}
%         \includegraphics[width=0.95\linewidth]{figs/add_pipeline.pdf}
%         \caption{Subject Addition Dataset Construction Pipeline}
%         \label{fig:add_pipeline}
%     \end{subfigure}
%     % 第二个子图
%     \begin{subfigure}{0.45\textwidth}
%         \includegraphics[width=0.88\linewidth]{figs/replace_pipeline.pdf}
%         \caption{Subject Replacement Dataset Construction Pipeline}
%         \label{fig:replace_pipeline}
%     \end{subfigure}

%     \caption{Data construction pipelines for instruction-based subject-driven image editing.}
%     \label{fig:data_construction}
% \end{figure*}

\section{Experiments}
In this section, we present experimental results on three tasks, followed by analysis of the joint training strategy and framework design. For subject-driven image generation, we compare our method with both task-specific models using MLLMs as encoders and recent unified frameworks. For instruction-based image editing, we similarly compare against both task-specific and unified models, including SmartEdit \cite{huang2024smartedit} and ml-MGIE \cite{fu2023guiding}, which also use MLLMs. For instruction-based subject-driven image editing, given the lack of instruction-only baselines, we additionally evaluate mask-dependent methods using masks from MIGEBench.

\subsection{Implementation Details}
\method includes a conditional diffusion model and a multimodal encoder. Our design allows flexible selection of various diffusion models, and we initialize with PIXART-$\alpha$ \cite{chen2023pixart} pretrained at a 512×512 resolution. The multimodal encoder consists of a pretrained Flan-T5-XXL \cite{chung2024scaling} as LLM for initialization and an image encoding component. This includes query tokens, Q-Former, and a projector, all initialized with a BLIP-2 \cite{li2023blip} checkpoint (pretrain\_flant5xxl). The frozen VAE encoder, used as the visual feature extractor, is the same as the one in the diffusion model. During training, we sample instruction-based subject addition and subject replacement data in a 1:1 ratio. Separately, subject-driven editing and instruction-based editing data are also sampled in a 1:1 ratio. A training strategy is employed with a 5\% probability of dropping either the conditional input or the multimodal condition, and an additional 5\% chance of dropping both, enabling classifier-free guidance during inference.

\subsection{Evaluation Results}
As a unified model, \method demonstrates exceptional performance across various image generation and editing tasks, even outperforming existing task-specific models. \method achieves strong performance in subject-driven generation and instruction-based editing, and shows promising results on instruction-based subject-driven image generation. Detailed results are in Supplementary Material.%~\ref{sec:inplement}.

%%%%%%%%%%%%%%%%%%%%%%%%%%%%%%%%%%%%%%%%%%%
%% 编辑的表格
\setlength{\tabcolsep}{1.5pt} % 调整列间隔
\begin{table*}[t]
\centering
\caption{\textbf{Quantitative results on the Emu Edit and MagicBrush test sets.} \method outperforms other universal models on MagicBrush, demonstrating superior instruction alignment and detail preservation.}
\scalebox{0.68}{
\begin{tabular}{c l >{\centering\arraybackslash}p{1.47cm} >{\centering\arraybackslash}p{1.47cm} >{\centering\arraybackslash}p{1.47cm} >{\centering\arraybackslash}p{1.47cm} >{\centering\arraybackslash}p{1.47cm} >{\centering\arraybackslash}p{1.47cm}|>{\centering\arraybackslash}p{1.47cm} >{\centering\arraybackslash}p{1.47cm} >{\centering\arraybackslash}p{1.47cm} >{\centering\arraybackslash}p{1.47cm} >{\centering\arraybackslash}p{1.47cm} >{\centering\arraybackslash}p{1.47cm}} 
\toprule
& & \multicolumn{6}{c|}{Emu Edit Test set~} & \multicolumn{6}{c}{MagicBrush Test set} \\ 
\midrule
&
Methods & 
DINO $\uparrow$ & 
CLIP-I $\uparrow$ & 
CLIP-T $\uparrow$ & 
CLIP$_{dir}\uparrow$ & 
L1 $\downarrow$ & 
L2 $\downarrow$ & 
DINO $\uparrow$ & 
CLIP-I $\uparrow$ & 
CLIP-T $\uparrow$ & 
CLIP$_{dir}\uparrow$ & 
L1 $\downarrow$ & 
L2 $\downarrow$ \\
% \midrule
% \multicolumn{1}{c}{} & \multicolumn{12}{c}{Task-specific Models} \\
\cmidrule{1-14}
\multirow{5}{*}{\makecell{Task-specific \\ Models}} 
& InstructPix2Pix \cite{brooks2023instructpix2pix} & 0.759 & 0.831 & 0.288 & 0.086 & 0.122 & 0.036 & 0.763 & 0.843 & 0.289 & 0.105 & 0.097 & 0.028 \\
& MagicBrush \cite{zhang2024magicbrush} & 0.800 & 0.857 & 0.295 & 0.102 & 0.085 & \underline{0.027} & 0.847 & 0.888 & 0.304 & \textbf{0.127} & 0.062 & 0.018 \\
& UltraEdit \cite{zhao2024ultraedit} & \textbf{0.862} & \underline{0.875} & \underline{0.301} & 0.100 & \textbf{0.050} & \textbf{0.008} & \underline{0.879} & 0.897 & \underline{0.305} & \underline{0.119} & \textbf{0.042} & \textbf{0.006} \\
& SmartEdit \cite{huang2024smartedit} & 0.752 & 0.830 & 0.290 & 0.100 & 0.129 & 0.055 & 0.821 & 0.879 & 0.301 & 0.122 & 0.084 & 0.033 \\
& ml-MGIE \cite{fu2023guiding} & 0.779 & 0.843 & 0.276 & 0.040 & 0.101 & 0.034 & 0.751 & 0.844 & 0.274 & 0.039 & 0.102 & 0.034 \\
% \midrule
% \multicolumn{1}{c}{} & \multicolumn{12}{c}{Unified Models} \\
\midrule
\multirow{8}{*}{\makecell{Unified\\Models}}
& PixWizard \cite{lin2024pixwizard} & 0.824 & 0.862 & 0.288 & 0.066 & 0.105 & 0.038 & 0.865 & 0.897 & 0.289 & 0.054 & 0.073 & 0.023 \\
& ACE \cite{han2024ace} & \underline{0.847} & \textbf{0.877} & 0.299 & 0.104 & \underline{0.079} & 0.028 & 0.856 & \underline{0.899} & 0.304 & \textbf{0.127} & 0.065 & 0.023 \\
& OmniGen \cite{xiao2024omnigen} & 0.766 & 0.819 & 0.299 & \textbf{0.123} & 0.160 & 0.065 & 0.821 & 0.863 & 0.289 & 0.087 & 0.116 & 0.045 \\
& OneDiff \cite{le2024one} & 0.728 & 0.819 & 0.285 & 0.082 & 0.136 & 0.060 & 0.742 & 0.840 & 0.295 & 0.107 & 0.128 & 0.052 \\
& \textbf{MIGE (ours)} & 0.832 & 0.865 & \textbf{0.306} & \underline{0.114} & 0.088 & \underline{0.027} & \textbf{0.889} & \textbf{0.905} & \textbf{0.306} & \underline{0.119} & \underline{0.055} & \underline{0.013} \\
&/only\_edit data & 0.785 & 0.841 & 0.302 & 0.104 & 0.117 & 0.046 & 0.796 & 0.862 & 0.300 & 0.111 & 0.094 & 0.036 \\
&/wo\_VAE feature & 0.799 & 0.846 & 0.300 & 0.098 & 0.103 & 0.035 & 0.811 & 0.868 & 0.299 & 0.105 & 0.098 & 0.036 \\
&/wo\_multimodal instruction & 0.592 & 0.723 & 0.280 & 0.113 & 0.177 & 0.074 & 0.548 & 0.714 & 0.277 & 0.135 & 0.170 & 0.066 \\
\bottomrule
\end{tabular}
}
\label{tb:editing}
\end{table*}

%%%%%%%%%%%%%%%%%%%%%%%%%%%%%%%%%%
\subsubsection{Instruction-based Image Editing}
Instruction-based image editing allows users to modify a source image using free-form multimodal instructions, such as adding, removing, or altering objects and styles. Table~\ref{tb:editing} reports results on Emu Edit and MagicBrush~\cite{zhang2024magicbrush}. DINO and CLIP-I assess similarity to the source image, CLIP-T evaluates alignment with the target caption, CLIP$_{dir}$ measures consistency between text and image embedding changes, and L1/L2 capture pixel-level differences.

\begin{figure}[htbp]
    \centering
    \includegraphics[width=0.44\textwidth]{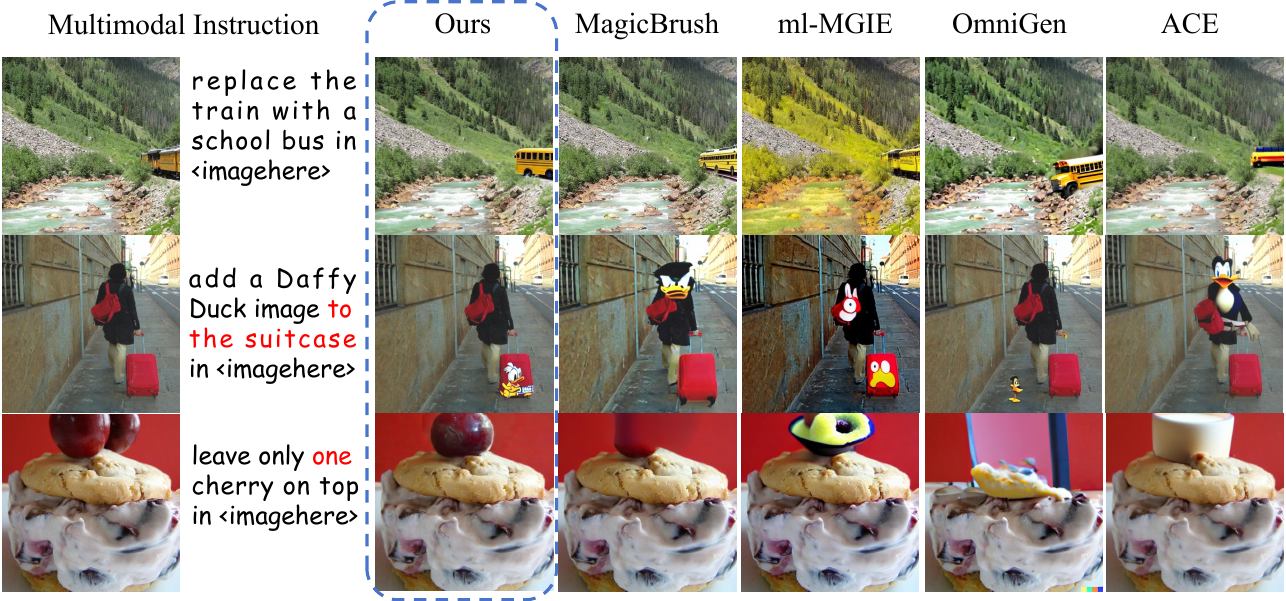} 
    \caption{Qualitative comparison for instruction-based image editing. MIGE demonstrates superior editing accuracy and instruction understanding.}
    \label{fig:compare_edit}
\end{figure}

As shown in Table~\ref{tb:editing}, \method achieves the highest CLIP-T and outperforms all task-specific models on CLIP$_{dir}$, indicating stronger instruction following. On MagicBrush, it surpasses all universal models across almost all metrics, with the best DINO, CLIP-I, CLIP-T, and lowest L1/L2, demonstrating superior instruction fidelity and detail preservation. This capability is further illustrated in Figure \ref{fig:compare_edit}, where \method is the only model that accurately follows the instruction to add a Daffy Duck image to the red suitcase without altering other unrelated areas. Compared to ml-MGIE, which also uses a MLLM as an encoder, MIGE better preserves input-output consistency and avoids altering irrelevant backgrounds.
%%%%%%%%%%%%%%%%%%%%%%%%%%%%%%%%%%%

%%%%%%%%%%%%%%%%%

\subsubsection{Subject-driven Image Generation}
\begin{figure}[htbp]
    \centering
    \includegraphics[width=0.4\textwidth]{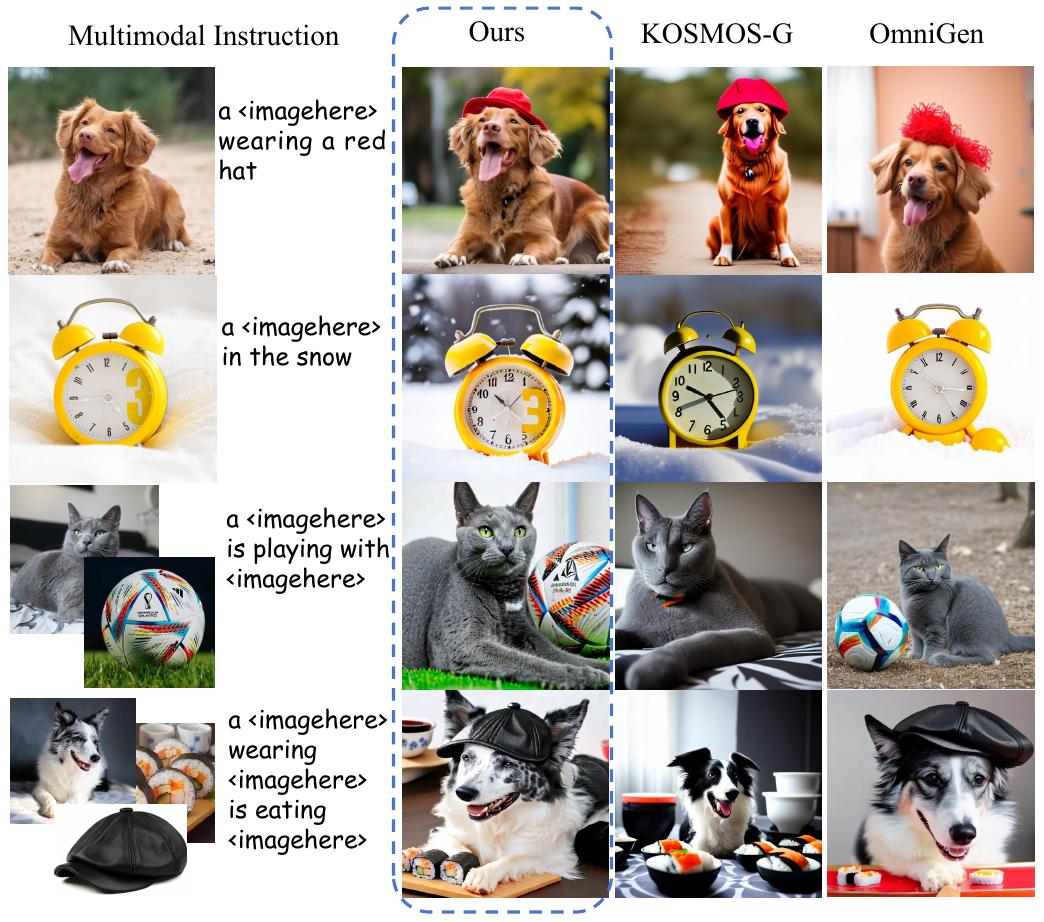} 
    \caption{Qualitative comparison for subject-driven image generation. MIGE exhibits superior subject preservation, even in multi-subject scenarios.}
    \label{fig:compare_sub}
\end{figure}

Generating images that satisfy both image and text constraints from a multimodal prompt is challenging. We compare \method with three task-specific methods using MLLMs as encoders and three universal models on DreamBench \cite{ruiz2023dreambooth}. DINO and CLIP-I measure subject fidelity, while CLIP-T assesses instruction adherence. As shown in Table~\ref{tb:dreambooth}, \method surpasses all unified models in subject preservation ability, as evidenced by the highest DINO and CLIP-I scores, with the DINO score even outperforming task-specific models. Examples in Figure~\ref{fig:compare_sub} further highlight \method's ability to effectively preserve the identity of each object in multi-object scenarios and combine them based on instructions, demonstrating that our multimodal encoder effectively captures both semantic and visual features simultaneously.
%excels in both single- and multi-subject generation, maintaining each subject’s distinct features, whereas other models often miss or blur them. This advantage stems from its ability to flexibly incorporate multiple reference entities and fuse visual features effectively.

%%%%%%%%%%%%%%%%%%%%%%%%%%%%%%%%%%%%%%%%%%%%%%%%%%%%%%%%%
%% 主体保持的表格
\begin{table}[h]
\setlength{\tabcolsep}{10pt}
\centering
\caption{Quantitative results for subject-driven image generation on DreamBench. Results with * denote our reproduction. MIGE outperforms universal models in subject preservation and is competitive with task-specific models.}
\scalebox{0.75}{
\begin{tabular}{clccc} 
\toprule
& Methods & DINO $\uparrow$ & CLIP-I $\uparrow$ & CLIP-T $\uparrow$ \\ 
\midrule
%\multicolumn{5}{c}{Task-specific Models} \\ 
%\midrule
\multirow{3}{*}{\makecell{Task-specific\\Models}}
& BLIP-Diffusion \cite{li2024blip} & 0.670 & 0.805 & 0.302 \\
& KOSMOS-G \cite{pan2023kosmos} & 0.694 & \textbf{0.847} & 0.287 \\
& UNIMO-G \cite{li-etal-2024-unimo} & 0.668 & \underline{0.841} & \textbf{0.329} \\
\midrule
%\multicolumn{4}{c}{Unified Models} \\ 
%\midrule
\multirow{6}{*}{\makecell{Unified\\Models}}
& Omnigen * \cite{xiao2024omnigen} & \underline{0.711} & 0.800 & 0.312 \\
& UniReal \cite{chen2024unireal} & 0.702 & 0.806 & \underline{0.326} \\
& OneDiffusion * \cite{le2024one} & 0.582 & 0.750 & 0.240 \\
& \textbf{MIGE (ours)} & \textbf{0.744} & 0.830 & 0.293 \\
& /only\_subject data & 0.726 & 0.823 & 0.289 \\
& /wo\_VAE feature & 0.741 & 0.828 & 0.287 \\
\bottomrule
\end{tabular}
}
\label{tb:dreambooth}
\end{table}
%%%%%%%%%%%%%%%%%%%%%%%%%
%%%%%%%%%%%%%%%%%%%%%%%%%%%%%%%%%%%%%

\begin{table*}[t]
\centering
\caption{\textbf{Quantitative results on instruction-based subject-driven editing.} Methods marked with a cross in the instruction column use masks, while the others generate images using multimodal instructions. Overall, MIGE significantly outperforms others in both tasks, showcasing superior editing and subject preservation abilities.}
\label{tb:new_task}
% 第一个子表格 (subject replacement)
\begin{subtable}[t]{0.492\textwidth}
\centering
\caption{Results on instruction-based subject replacement.}
\scalebox{0.7}{
\begin{tabular}{clccclcc} 
\toprule
&\multicolumn{1}{c}{Methods} & \multicolumn{3}{c}{Editing} &  & \multicolumn{2}{c}{Subject Preserving} \\ 
\cline{3-5}\cline{7-8}
& & DINO $\uparrow$ & CLIP-I $\uparrow$ & CLIP-T $\uparrow$ &  & DINO $\uparrow$ & CLIP-I $\uparrow$ \\ 
\midrule
&\textit{I-O sim} & 0.668 & 0.842 & 0.271 &  &  &  \\ 
\midrule
\multirow{4}{*}{\makecell{Mask-denpendent\\Models}}
% \multicolumn{1}{c}{} & \multicolumn{6}{c}{Mask-denpendent methods} \\
% \midrule
&PBE \cite{yang2023paint} & 0.810 & 0.885 & 0.304 &  & 0.521 & 0.792 \\
&AnyDoor \cite{chen2024anydoor} & 0.817 & 0.890 & 0.307 & & 0.551 & 0.789 \\
&TIGIC \cite{li2025tuning} & 0.789 & 0.874 & \textbf{0.313} &  & 0.453 & 0.744 \\
&MADD \cite{he2024affordance} & 0.736 & 0.852 & 0.284 &  & 0.446 & 0.742 \\
\midrule
% \multicolumn{1}{c}{} & \multicolumn{6}{c}{Mask-free Models} \\
% \midrule
\multirow{5}{*}{\makecell{Mask-free\\Models}}
&OmniGen \cite{xiao2024omnigen}  & 0.802 & 0.868 & 0.296 &  & 0.580 & 0.792 \\
&\textbf{MIGE(ours)} & \textbf{0.863} & \textbf{0.909} & \underline{0.307} &  & \textbf{0.652} & \textbf{0.834} \\ 
&/subject data + edit data & 0.789 & 0.873 & 0.299 &  & 0.503 & 0.764 \\
&/only\_compositional data & 0.780  & 0.860 & 0.303 & & 0.585 & 0.808   \\
&/wo\_VAE feature & 0.852  & 0.900 & 0.308 & & 0.633 & 0.819   \\
\bottomrule
\end{tabular}
}

\label{tb:replace}
\end{subtable}
\hfill
% 第二个子表格 (subject addition)
\begin{subtable}[t]{0.492\textwidth}
\centering
\caption{Results on instruction-based subject addition.}
\scalebox{0.7}{
\begin{tabular}{clccclcc} 
\toprule
&\multicolumn{1}{c}{Methods} & \multicolumn{3}{c}{Editing} &  & \multicolumn{2}{c}{Subject Preserving} \\ 
\cline{3-5}\cline{7-8}

& & DINO $\uparrow$ & CLIP-I $\uparrow$ & CLIP-T $\uparrow$ &  & DINO $\uparrow$ & CLIP-I $\uparrow$ \\ 
\midrule
&\textit{I-O sim} & 0.783 & 0.903 & 0.295 &  &  &  \\
%\midrule
%\multicolumn{1}{c}{} & \multicolumn{6}{c}{Mask-denpendent Models} \\
\midrule
\multirow{4}{*}{\makecell{Mask-denpendent\\Models}}
&PBE \cite{yang2023paint} & 0.843 & 0.908 & 0.321 &  & 0.495 & 0.794 \\
&AnyDoor \cite{chen2024anydoor} & 0.863 & 0.930 & 0.324 & & 0.533 & 0.799 \\
&TIGIC \cite{li2025tuning} & 0.840 & 0.901 & \textbf{0.325} &  & 0.455 & 0.753 \\
&MADD \cite{he2024affordance}  & \underline{0.885} & \underline{0.930} & 0.316 &  & 0.519 & 0.785 \\
%\midrule
%\multicolumn{1}{c}{} & \multicolumn{6}{c}{Mask-free methods} \\
\midrule
\multirow{5}{*}{\makecell{Mask-free\\Models}}
& OmniGen \cite{xiao2024omnigen}  & 0.791 & 0.870 & 0.312 &  & \underline{0.605} & \underline{0.814} \\
&\textbf{MIGE(ours)} & \textbf{0.909} & \textbf{0.940} & \underline{0.322} &  & \textbf{0.638} & \textbf{0.838} \\ 
&/subject data + edit data & 0.807 & 0.895 & 0.304 &  & 0.309 & 0.683 \\ 
&/only\_compositional data & 0.879  & 0.928 & 0.324 & & 0.577 & 0.820   \\
&/wo\_VAE feature  & 0.902  & 0.934 & 0.322 & & 0.633 & 0.831   \\
\bottomrule
\end{tabular}
}
\label{tb:add}
\end{subtable}
\end{table*}

%%%%%%%%%%%%%%%%%%%%%%%%%%%%%%
\subsubsection{Instruction-based Subject-driven Image Editing}
\paragraph{Benchmark Construction}
Instruction-based subject-driven image editing is a novel task. Existing methods rely on masks or positional cues for editing \cite{li2023dreamedit,yang2024dreammix} but lack support for instruction-only editing. The AnyEdit-Test Benchmark \cite{yu2024anyedit} supports few instruction templates and evaluates only one type of object editing based on automatically extracted masks. Current benchmarks \cite{yang2023paint,chen2024anydoor} for subject addition and replacement separately evaluate foreground and background similarities without providing a complete edited image as ground truth, making them unsuitable for this task.

To address these issues, we manual construct MIGEBench, a benchmark consisting of 1,000 test samples: 500 for instruction-based subject addition and 500 for subject replacement. More details can be found in the Supplementary Material.
%We select editing data, extract the changed entities, and create multimodal instructions and target captions. To ensure quality, we perform additional  filtering. 
%using SEED-Data-Edit \cite{ge2024seed}. Subjects are extracted using Grounded SAM \cite{ren2024grounded}, while Qwen2.5-14B-Instruct \cite{qwen2.5} extracts entity names and constructs multimodal instructions. 我Target captions are then generated by GPT-4o and manually refined. To ensure compatibility, masks are also provided. More details can be found in Supplementary Material. %~\ref{sec:appen_benchmark_construction}.
\\
\paragraph{Evaluation Results}
Our evaluation focuses on editing ability and subject preservation. DINO and CLIP-I measure the similarity of the generated image to the ground truth, while CLIP-T evaluates alignment with the target caption, indicating editing ability. Subject preservation is evaluated by extracting the edited subject via Grounded SAM \cite{ren2024grounded} and comparing it to the input subject image using DINO and CLIP-I. This separates the evaluation of image-level editing from subject-level feature preservation. Methods that do not support instruction-based editing use masks provided in MIGEBench during testing.

\begin{figure*}[htbp]
    \centering
    \includegraphics[width=0.7\textwidth,height=6.3cm]{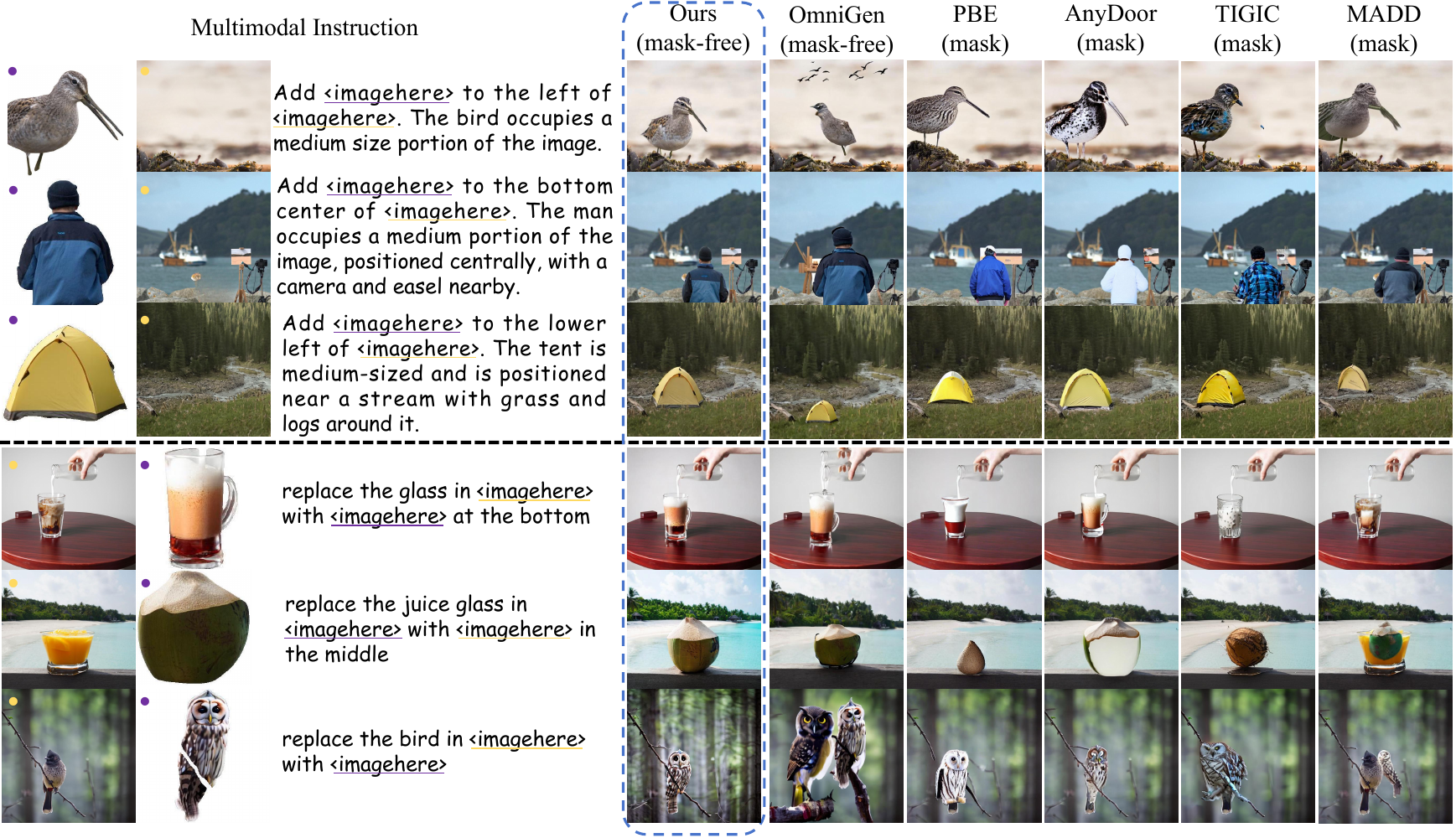} 
    \caption{\textbf{Qualitative results on the benchmark for the subject addition and subject replacement.} The upper section compares subject addition results, while the lower section compares subject replacement. During testing, the <imagehere> placeholder in the multimodal instruction is replaced according to the image sequence. MIGE demonstrates flexibility in editing and excels in subject preservation ability and input-output consistency.}
    \label{fig:compare_new}
\end{figure*}

Quantitative comparisons with other methods are shown in Tables \ref{tb:replace} and \ref{tb:add}. We compute the DINO, CLIP-I, and CLIP-T scores between the input image and the ground truth. The first row of the table (denoted as "I-O sim") provides a baseline for evaluating the effectiveness of editing ability. Mask-dependent models can easily preserve the edit-unrelated regions, but \method still outperforms all methods, highlighting the effectiveness of conditional inputs and \method's precise understanding of instructions. The CLIP-T score measures instruction adherence in mask-free models. \method significantly outperforms OmniGen in both tasks, demonstrating the advantage of our unified approach. \method also achieves the best performance on subject preservation in both tasks. As shown in Figure~\ref{fig:compare_new}, \method successfully understands the task of replacing a bird with an owl, rather than simply overlaying or adding an owl, demonstrating the effectiveness of the multimodal encoder which simultaneously encoding the semantic and visual information of the reference image.

%%%%%%%%%%%%%%%%%%%%%%%%%%%%%%%%%%%%%%%%%%%%%%%%%%%%%

\subsection{Further Analysis}
\subsubsection{Effectiveness of Joint Training}
To assess the effectiveness of joint training, we train models separately on individual datasets (denoted as ``only\_subject data,'' ``only\_edit data,'' and ``only\_compositional data'') and compare their performance with the jointly trained model. The results in Tables \ref{tb:editing}, \ref{tb:dreambooth} and \ref{tb:new_task} show that joint training leads to consistent improvements across all metrics, demonstrating that subject-driven generation and instruction-based editing reinforce each other. As shown in Table \ref{tb:new_task}, joint training also enhances the performance of the compositional new task, further highlighting its overall benefits. These findings emphasize both the effectiveness and necessity of joint training. In conclusion, joint training of subject-driven generation and instruction-based editing within our unified framework not only boosts compositional capability but also improves the performance of each individual task. This validates our motivation and provides further theoretical support for unified models for multiple tasks.

\subsubsection{Effectiveness of Feature Fusing}
\method employs a feature fusion mechanism in the multimodal encoder to integrate semantic features from ViT and visual features from VAE. As shown in Table \ref{tb:dreambooth}, Table \ref{tb:editing} and Table \ref{tb:new_task}, compared to the model without VAE features (denoted as ``wo\_VAE feature''), incorporating VAE features significantly improves detail preservation in reference images, benefiting both subject-driven image generation and instruction-based image editing. This is particularly evident in the improved CLIP-I and DINO scores and the significant reduction in L1 and L2 metrics, demonstrating that the inclusion of additional visual features helps maintain consistency between the input and output.

\subsubsection{Effectiveness of Instruction-based Subject-driven Image Editing Dataset}
Joint training of subject-driven image generation and instruction-based image editing enables generalization to instruction-based subject-driven image editing (denoted as ``subject data + edit data''). To enhance the capability of \method in this new task, particularly in understanding spatial terms and size descriptions, we constructed a task-specific dataset for joint training. As shown in Table \ref{tb:replace} and Table \ref{tb:add}, the task-specific data significantly improved the model’s overall abilities. This demonstrates the effectiveness of our constructed dataset, and the proposed data generation pipeline serves as a valuable reference for future dataset construction.

\subsubsection{Effectiveness of Multimodal Instruction}
Existing instruction-based editing works \cite{brooks2023instructpix2pix,sheynin2024emu,zhang2024magicbrush} typically use text instructions as the conditional input, while we extend this to multimodal instructions. To measure the benefit of multimodal instructions, we trained the model with text-only editing instructions for comparison. As a unified model, \method demonstrates exceptional performance across various image generation and editing tasks, even outperforming existing task-specific models. As shown in Table \ref{tb:editing}, using multimodal instructions consistently improves performance over text-only instructions (denoted as ``wo\_multimodal instruction''). This enhances input-output consistency and instruction-following ability in multi-task training. The significant improvement in the L1 and L2 metrics indicates finer control over images and more accurate edits. While text-only instructions provide the necessary changes, the high CLIP$_{dir}$ score and lower values in other metrics show that multimodal instructions add visual context, enabling more precise and faithful modifications. For more qualitative results, see Supplementary Material.%~\ref{sec:qr}.

\subsubsection{Comparison of Data Volume}
Task-specific models treat tasks separately and typically require large amounts of high-quality data. In contrast, by exploiting the relationships between tasks and the unified structure, \method achieves excellent results with minimal data through joint training. As shown in Table \ref{tb:data_volume}, \method demonstrates that both the model structure and joint training are key to achieving this performance with minimal data, using fewer data than task-specific models and other unified models.
%%%%%%%%%%%%%%%%%%%%%%%%%%%
\begin{table}[h]
\setlength{\tabcolsep}{20pt}
\centering
\caption{Comparison of Data Volume Across Methods}
\scalebox{0.75}{
\begin{tabular}{clc} 
\toprule
& Methods & Data Volume\\ 
\midrule
%\multicolumn{5}{c}{Task-specific Models} \\ 
%\midrule
\multirow{4}{*}{\makecell{Task-specific\\Models}}
& MADD  \cite{he2024affordance} & 3M  \\
& PBE  \cite{yang2023paint} & 16M  \\
& KOSMOS-G \cite{pan2023kosmos} & 3159.3M  \\
& UNIMO-G \cite{li-etal-2024-unimo} & 304.3M  \\
\midrule
%\multicolumn{4}{c}{Unified Models} \\ 
%\midrule
\multirow{5}{*}{\makecell{Unified\\Models}}
& PixWizard \cite{lin2024pixwizard} & 30M \\
& UniReal \cite{chen2024unireal} & 100M \\
& ACE \cite{han2024ace} & 700M   \\
& Omnigen \cite{xiao2024omnigen} & 326.06M \\
& \textbf{MIGE (ours)} & \textbf{2.28M} \\
\bottomrule
\end{tabular}
}
\label{tb:data_volume}
\end{table}

%However, this results in limited generalization capability and necessitates extensive training data.

\section{Conclusion}
We present \method, a unified framework that combines subject-driven generation and instruction-based editing, leveraging multimodal instructions and conditional input for joint training that enhances task synergy. Joint training also unlocks new capabilities like instruction-based subject-driven image editing. We introduce pipelines for this new task to construct training data and MIGEBench for evaluation.
Our experiments show that joint training leads to significant enhancement in subject fidelity and instruction adherence, demonstrating the effectiveness of unifying these tasks. This integration enhances controllability and offers promising directions for future multimodal image generation and editing.

%\section{Supplementary Material}

% \begin{acks}
% \end{acks}

\clearpage
\section{Supplementary Material}
%\afterpage{\clearpage}
\appendix

\section{Qualitative Comparison}
\label{sec:qr}
\begin{figure*}[h]
    \centering
    \includegraphics[width=0.9\textwidth]{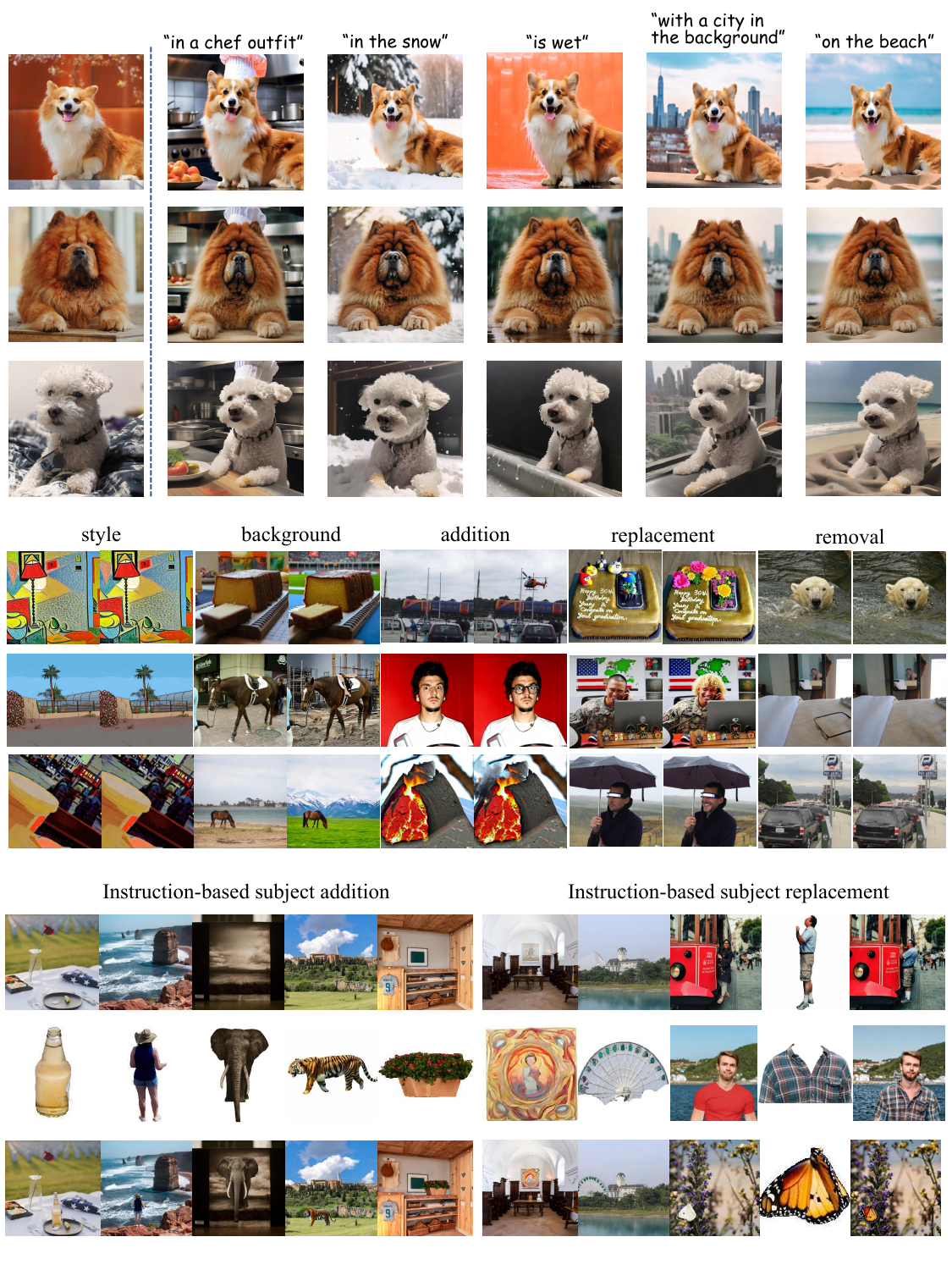} 
    \caption{Qualitative results of subject-driven image generation (top) , instruction-based image editing (middle), and instruction-based subject-driven image editing (bottom).}
    \label{fig:qualitative}
\end{figure*}

We conduct a qualitative comparison to assess the visual performance and effectiveness of \method. 
%Figure \ref{fig:compare_main} and Figure \ref{fig:compare_new} compares the performance of \method across three tasks with other methods, focusing on subject fidelity, instruction adherence, and overall image quality. 
Figure \ref{fig:qualitative} presents a range of cases on each task to showcase the generation results of \method, demonstrating its ability to handle diverse inputs and produce high-quality outputs. This comparison highlights the strengths of \method and the benefit of joint training, offering valuable insights into its potential for real-world applications.

\section{Related Work}

\label{sec:related work}
\subsection{Subject-Driven Image Generation}
Subject-driven generation focuses on preserving given subject features while generating new images. DreamBooth \cite{ruiz2023dreambooth} fine-tunes the full model, leading to high resource costs. IP-Adapter \cite{ye2023ip} trains lightweight attention layers, struggles with prompt-image balance. Methods like Blip-Diffusion \cite{li2024blip} use embeddings to condition generation, yet face challenges in subject preservation and generalization. KOSMOS-G \cite{pan2023kosmos} and UNIMO-G \cite{li-etal-2024-unimo} leverage Multimodal Large Language Models  (MLLMs) as flexible encoders, but require large datasets for task-specific adaptation and lack robust instruction-following ability, limiting transferability.

\subsection{Instruction-based Image Editing}
Instruction-based editing modifies images based on instructions, offering greater flexibility than caption-driven methods. Inst-inpaint \cite{yildirim2023inst} and EraseDraw \cite{canberk2025erasedraw} focus on object-level edits. InstructPix2Pix \cite{brooks2023instructpix2pix}, EmuEdit \cite{sheynin2024emu}, and MagicBrush \cite{zhang2024magicbrush} enable diverse edits via large-scale data. AnyEdit \cite{yu2024anyedit}, FireEdit \cite{zhou2025fireedit}, and TF-TI2I \cite{hsiao2025tf} introduce task-aware routing, region prompts, or fixed templates. Recent works like InsightEdit \cite{xu2024insightedit}, MGIE \cite{fu2023guiding}, and SmartEdit \cite{huang2024smartedit} leverage MLLMs to enhance instruction understanding. However, maintaining input-output consistency remains challenging.

\section{Dataset Construction}
\label{sec:appen_data construction}
\begin{figure}[htbp]
    \centering
    \includegraphics[width=0.25\textwidth]{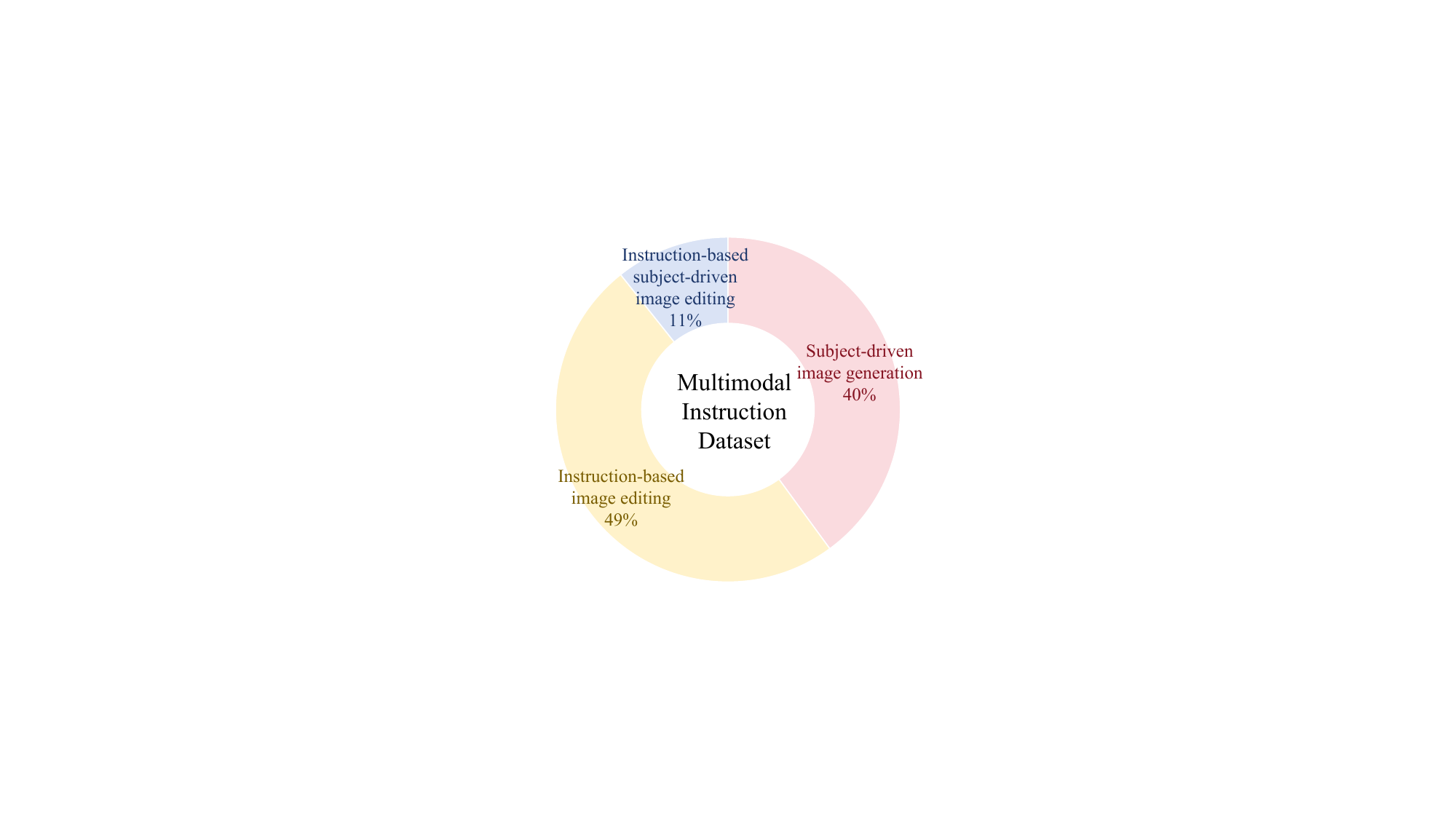} 
    \caption{Composition of training data.}
    \label{fig:data_ratio}
\end{figure}

To enable joint training of \method, we designed a series of pipelines for data construction and processing. Our dataset includes three tasks: subject-driven image generation, instruction-based image editing, and instruction-based subject-driven image editing. The proportions of the training data are shown in Figure \ref{fig:data_ratio}. In this section, we will detail the pipelines for processing each data component and other related details.\\

\subsection{Subject-driven Image Generation Data Construction}
We construct multimodal instruction-image pairs by replacing entities in image captions with corresponding entity images. The dataset combines BLIP3-GROUNDING-50M \cite{xue2024xgen} and internal sources. For the former, we filter out low-quality pairs with CLIP-T scores below 0.255 and retain entities with bounding boxes sized between 0.05 and 0.8. Entities are segmented using SAM \cite{kirillov2023segment}, and pairs with low CLIP-T scores between the prompt "a photo of [entity]" and the segmented image are removed. For internal data, we adopt the KOSMOS-G's pipeline, using MPT-7B-Instruct \cite{MosaicML2023Introducing} to extract entities. Entity images are obtained via Grounded SAM \cite{ren2024grounded}, and filtered by a CLIP-T score threshold of 0.255 and a size ratio between 0.1 and 0.85. To enhance subject fidelity, we further include 112,846 samples from subject200k, following OminiControl \cite{tan2024ominicontrol}.
\\
\subsection{Instruction-based Image Editing Data Construction}
Previous works on instruction-based editing relied solely on text, covering object-level additions, deletions, modifications, and global changes in background or style. We integrated data from InstructPix2Pix \cite{brooks2023instructpix2pix}, UltraEdit \cite{zhao2024ultraedit}, MagicBrush \cite{zhang2024magicbrush}, SEED-Data-Edit \cite{ge2024seed}, GIER \cite{shi2020benchmark}, and HQ-Edit \cite{hui2024hq}, filtering out instances with target image aesthetic scores above 5.5. To enhance multimodal instruction alignment, we reformulated text-based instructions based on task type. Specifically, for instructions containing phrases like ``in the image'' or ``to the image,'' we replaced them with ``in<imagehere>'' or ``to<imagehere>.'' For others, we appended ``of<imagehere>'' for global edits or ``in<imagehere>'' for localized modifications.
\\
%%%%%%%%%%%%%%%%%%%%%%%%%%%%%%%%%%%%%%%%%%%%%%%%%%%%%%%%%%%%%%%%%%%%%%%%%%%%%%%%%
\subsection{Instruction-based Subject-driven Image Editing Data Construction}
For subject addition, inspired by SAM-FB \cite{he2024affordance}, we construct background images, foreground entities, and multimodal instructions, as shown in Figure~\ref{fig:add_pipeline}. We filter SA-1B \cite{kirillov2023segment} images with aesthetic scores above 5 and select entities with a foreground-to-image ratio between 0.1 and 0.5. Qwen2-VL-7B \cite{Qwen2-VL} verifies entity completeness. The selected entities serve as foregrounds, and backgrounds are inpainted using LAMA \cite{suvorov2022resolution}. Foreground-related text is extracted with Qwen2-VL-7B, and multimodal instructions are generated via GPT-4o \cite{hurst2024gpt}. Due to resource constraints, we process the first 500,000 SA-1B samples, yielding 193,247 multimodal instruction-foreground-background pairs.

For subject replacement, we construct foreground entities and multimodal instructions, as shown in Figure \ref{fig:replace_pipeline}. SEED-Data-Edit's part1-unsplash is filtered to retain replace-task samples with annotation match scores above 0.3 and aesthetic scores above 5.5. Qwen2.5-14B-Instruct \cite{qwen2.5} verifies the main entity. The target entity is replaced with \texttt{<imagehere>} to form the multimodal prompt. Grounded SAM segments entities based on annotations, resulting in 79,693 multimodal instruction-replace entity-source image-target image pairs. We also include 34,947 virtual try-on samples generated using IDM-VTON \cite{choi2024improving}.
%%%%%%%%%%%%%%%%%%
\section{Prompts}
\label{sec:prompts}
LLMs and MLLMs are essential to our data and benchmark construction, with all prompts listed in Table~\ref{tab:subject_completeness} to Table~\ref{tab:edit_instruction_subject_extraction}.
%%%%%%%%%%%%%%%%%%%%%%%%%%%%%%%%%%%%%%%%%%%%%%%
\section{Implementation Details}
\label{sec:inplement}
\subsection{Training Details}
\method includes a conditional diffusion model and a multimodal encoder. The diffusion model is initialized from the PIXART-$\alpha$ checkpoint pretrained at a 512×512 resolution. Parameters introduced for handling conditional inputs are initialized to zero. The multimodal encoder consists of a pretrained Flan-T5-XXL \cite{chung2024scaling} as the LLM for initialization, and an image encoding component. This image encoder includes query tokens, Q-Former, and a projector, all initialized from the BLIP-2 \cite{li2023blip} checkpoint, with a frozen pretrained CLIP. The frozen VAE encoder, which acts as the visual feature extractor, is the same as the one in the diffusion model. Additionally, a zero-initialized MLP layer is introduced in the feature fusing mechanism for progressive visual feature integration. 

\method is trained on our multi-task dataset using the AdamW optimizer \cite{loshchilov2017decoupled} with a weight decay of 0.03 and a learning rate of 1e-5 for 18 epochs on 48 H20 GPUs, totaling six days of training with a batch size of 960 (20 per GPU).

%Our framework enables joint training through multimodal instruction and structural input unification. A conditional dropout strategy is applied, with a 5\% probability of dropping either the conditional input or the multimodal condition separately and an additional 5\% probability of dropping both, supporting classifier-free guidance during inference.
\subsection{Evaluation Details}
We evaluated the subject-driven generation capability of \method on DreamBench, which contains 750 prompts covering 30 subjects. For each prompt, four images were generated using seeds 0, 1, 2, and 3, resulting in 3,000 images. Following KOSMOS-G, we selected one image per prompt. Then we use the extracted subject from Grounded SAM as input, aligning with the training process. Subject fidelity was assessed using DINO and CLIP-I, while CLIP-T measured adherence to multimodal instructions.

For instruction-based editing evaluation, the Emu Edit benchmark contains known issues, including incorrect image-caption pairs and duplicate source-target captions. Prior works have handled these inconsistencies differently, leading to incomparable results. Therefore, we reimplemented all currently available open-source methods on both test sets using a fixed random seed (seed=0) for consistency. Consequently, CLIP-I, L1, and DINO were computed on 3,589 Emu Edit test samples, while CLIP$_{dir}$ and CLIP-T were evaluated on 2,899 samples after removing problematic data. 
%%%%%%%%%%%%%%%%%%%%%%%%

\begin{figure}[htbp]
    \centering
    \includegraphics[width=0.45\textwidth]{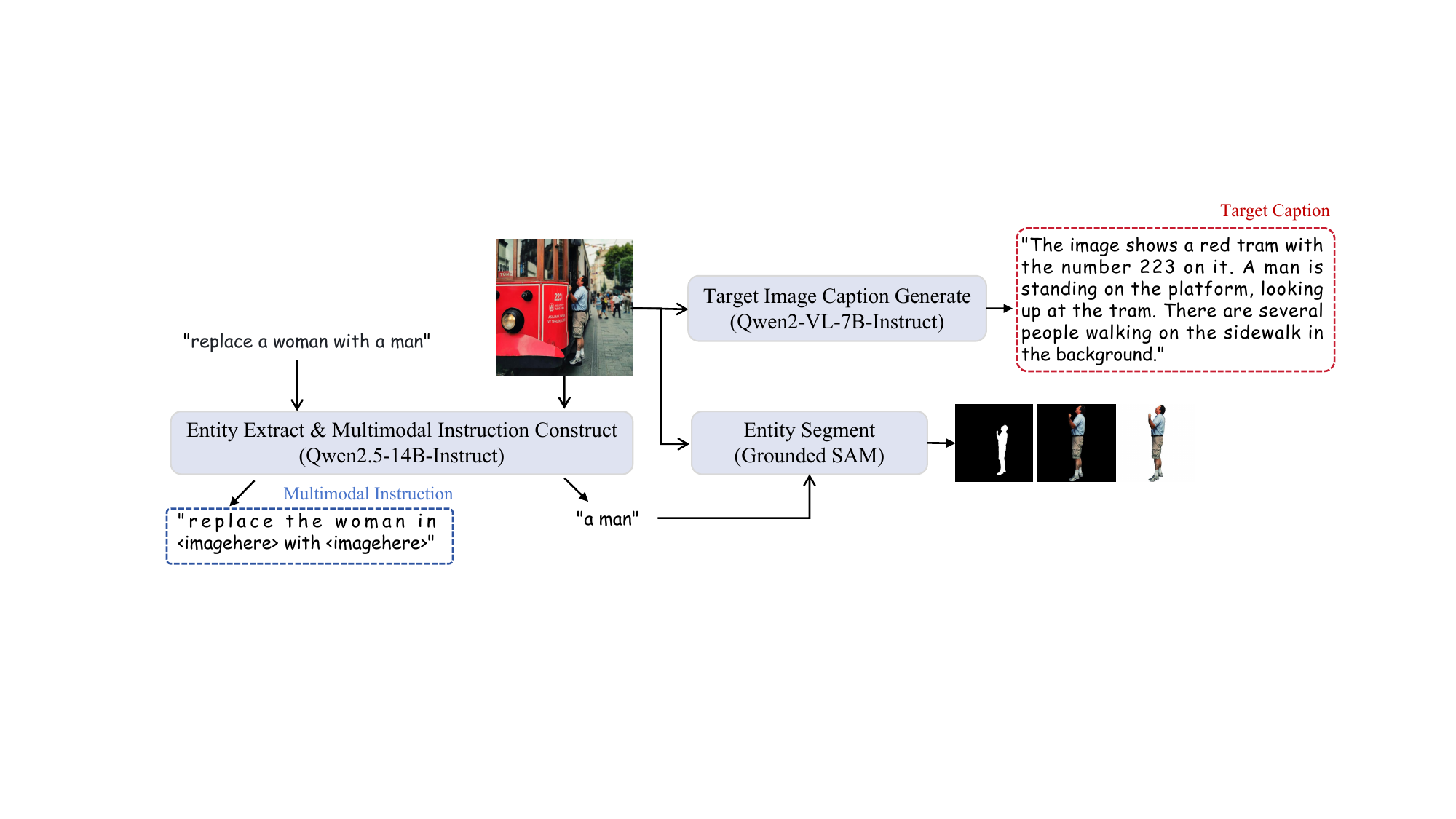} 
    \caption{The pipeline of benchmark construction.}
    \label{fig:benchmark_construct}
\end{figure}
%%%%%%%%%%%%%%%%%%%%%%%%%%%%%%%%%%%%%%%%%%%%%%%
\section{Benchmark Construction}
\label{sec:appen_benchmark_construction}
Instruction-based subject-driven image editing is a novel task lacking evaluation benchmarks. Previous works \cite{li2023dreamedit,yang2024dreammix} did not support variety of instructions and lacked ground truth target images. To address this, we designed a benchmark construction pipeline with multiple manual inspections to ensure image and caption quality.

\begin{figure}[htbp]
    \centering
    \includegraphics[width=0.3\textwidth]{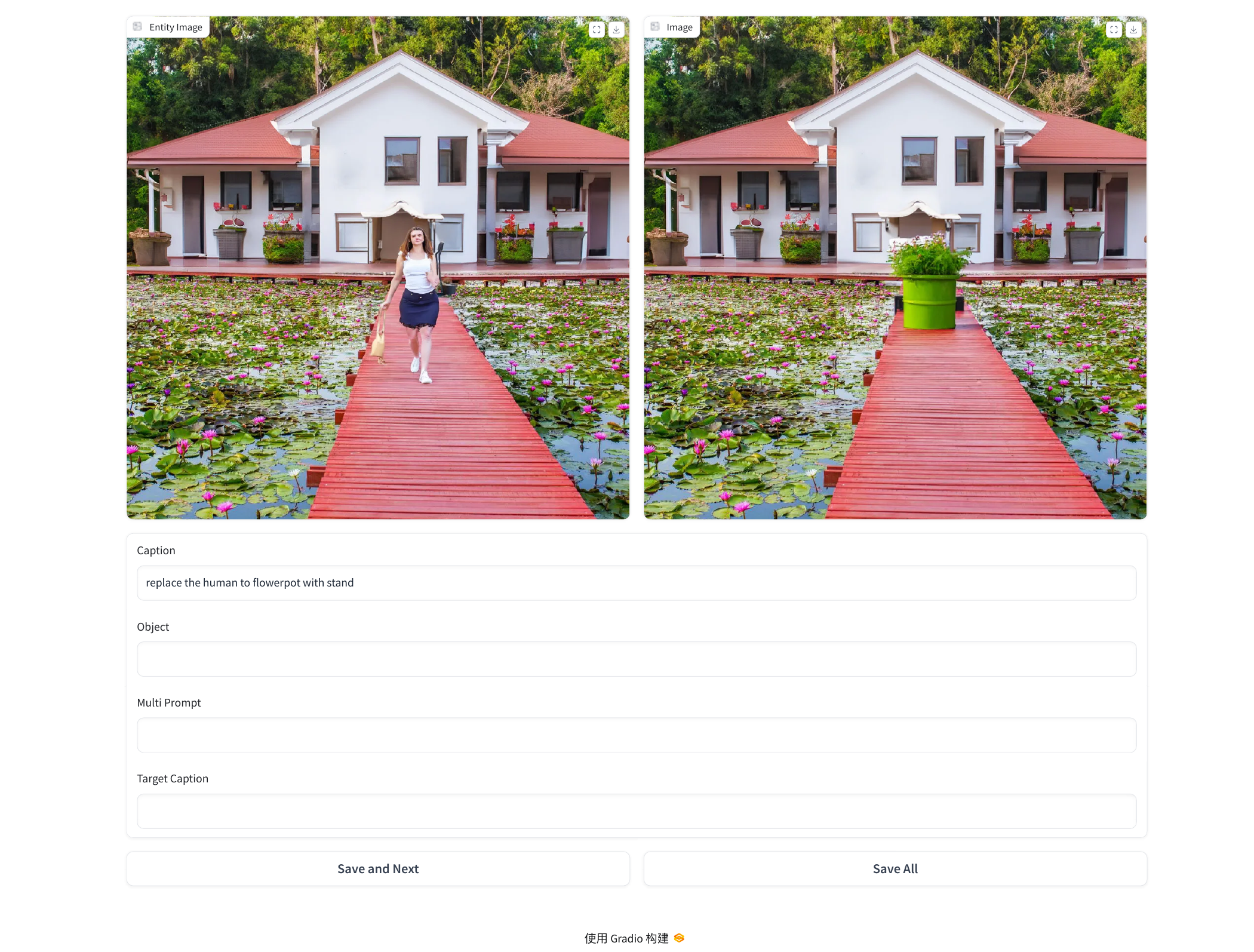} 
    \caption{The annotation interface for benchmarks built with Gradio.}
    \label{fig:gradio}
\end{figure}

As shown in Figure \ref{fig:benchmark_construct}, we first filter valid editing pairs using a Gradio \cite{abid2019gradio} interface, as shown in Figure \ref{fig:gradio}. Then, Qwen2.5-14B-Instruct extracts entity names and constructs multimodal instructions. After manually reviewing object names, we use Grounding DINO \cite{liu2025grounding} for bounding boxes and SAM for segmentation. Cropped entities are saved with black and white backgrounds for diverse test scenarios. Prompts are detailed in Supplementary Material \ref{sec:prompts}.

\begin{figure}[htbp]
    \centering
    \includegraphics[width=0.34\textwidth]{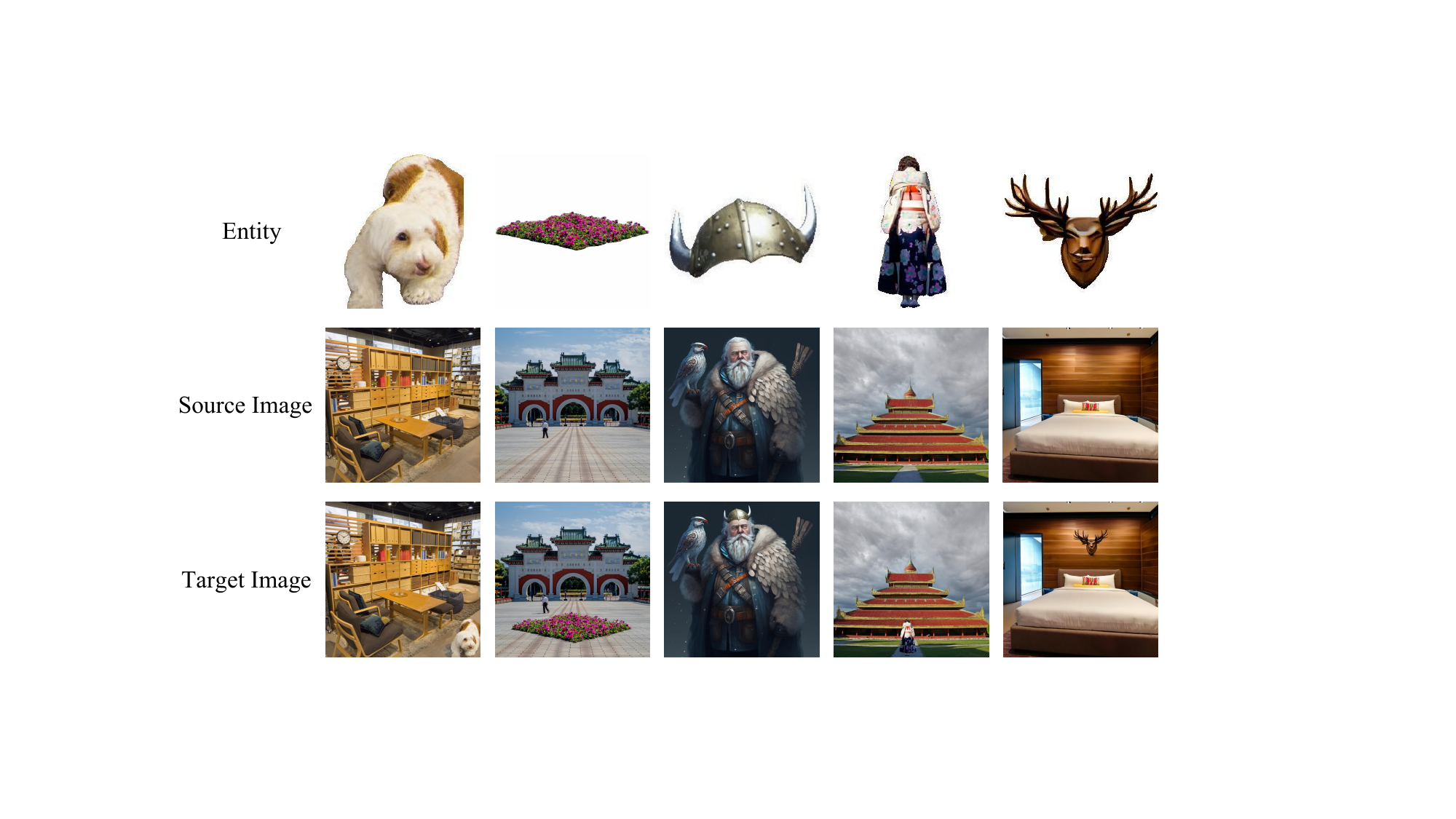} 
    \caption{Subject addition examples in our benchmark.}
    \label{fig:benchmark_add_case}
\end{figure}

\begin{figure}[htbp]
    \centering
    \includegraphics[width=0.34\textwidth]{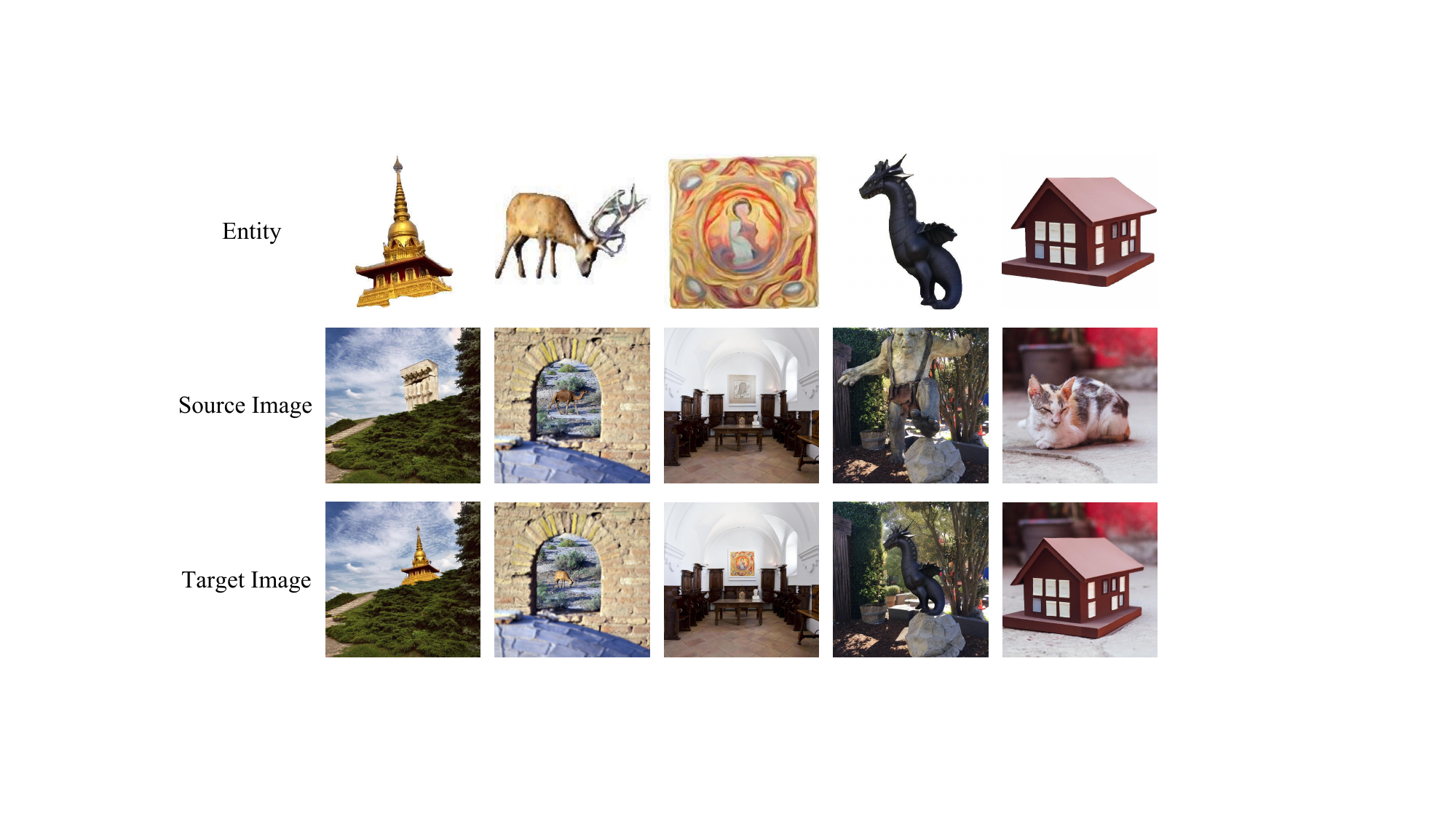} 
    \caption{Subject replacement examples in our benchmark.}
    \label{fig:benchmark_replace_case}
\end{figure}

For evaluation, Qwen2-VL-7B generates captions comparing generated and ground truth images, followed by manual reviews to remove unrecognized entities and correct inaccuracies. The final benchmark includes 500 samples each for subject addition and replacement. By avoiding artifacts from inpainting and incorporating multiple manual reviews, we ensure high-quality data. Figure \ref{fig:benchmark_add_case} and Figure \ref{fig:benchmark_replace_case} showcase some example benchmark pairs.

%%%%%%%%%%%%%%%%%

%%%%%%%%%%%%%%

%%%%%%%%%%%%%%%%%%%%%%%%%
Our benchmark, MIGEBench, is constructed based on SEED-Data-Edit, which is diversity. The benchmark covers a total of 565 distinct subjects. For the subject-replacement subset, the data is distributed across 10 major, well-balanced categories, as shown in \ref{tab:static}. This hierarchical and balanced structure ensures a high degree of diversity throughout the MIGEBench dataset.

\begin{table}[htbp]
\centering
\caption{Category-wise Statistics of MIGEBench}
\label{tab:static}
\scalebox{0.85}{ 
\begin{tabular}{lcc}
\toprule
\textbf{Category} & \textbf{Subject Replacement} & \textbf{Subject Addition} \\
\midrule
Animals & 62 & 77 \\
People & 60 & 66 \\
Vehicles & 40 & 52 \\
Buildings \& Structures & 27 & 50 \\
Plants \& Nature & 30 & 33 \\
Household \& Man-made Objects & 88 & 91 \\
Art \& Decor & 54 & 47 \\
Food \& Drink & 23 & 8 \\
Clothing \& Accessories & 31 & 24 \\
Other & 85 & 52 \\
\midrule
\textbf{Total} & \textbf{500} & \textbf{500} \\
\bottomrule
\end{tabular}
}
\end{table}

%%%%%%%%%%%%%%%%%%%%%%%%
%\clearpage
\begin{table*}[h]
\centering
\caption{Prompt design for subject completeness evaluation in an image. The model should identify whether the subject is complete, not abstract elements.}
\begin{tabular}{p{\textwidth}} 
\toprule
\rowcolor[gray]{0.9}
\textit{Prompt for Qwen2-VL-7B to determine the completeness of subject in the image} \\ 
\midrule
\textbf{Prompt:} Determine if the subject in the image is complete. If it is complete and not an abstract object such as background, grass, sky, tree, stone, or part of another item, please return \texttt{True}. Otherwise, return \texttt{False}.\\
\bottomrule
\end{tabular}
\label{tab:subject_completeness}
\end{table*}

\begin{table*}[h]
\centering
\caption{Prompt design for GPT-4o to generate multimodal instruction for subject addition training data.}
\begin{tabular}{p{\textwidth}} 
\toprule
\rowcolor[gray]{0.9}
\textit{Prompt for GPT-4o to generate multimodal instruction} \\ 
\midrule
\textbf{Prompt:} The object is \texttt{\{object\_name\}}. It is located in a bounding box with coordinates (\texttt{\{x\}}, \texttt{\{y\}}, \texttt{\{w\}}, \texttt{\{h\}}) on an image of size \texttt{\{width\}x\{height\}}. Describe its size, relative position, and relation to surrounding objects. Avoid describing the overall scene or unrelated elements.\\
Your response should start with ``Add \texttt{<imagehere>} to the \texttt{[position]} of \texttt{<imagehere>}''. (The first \texttt{<imagehere>} indicates the object and the second indicates the image.) Keep the \texttt{<imagehere>} symbol in the first sentence in your reply.\\
Answer briefly in two sentences:\\
\bottomrule
\end{tabular}
\label{tab:gpt-4o}
\end{table*}

\begin{table*}[h]
\centering
\caption{Prompt design for determining if an edit instruction pertains to the main subject of the image. The model evaluates whether the instruction relates to humans, animals, or other environmental features.}
\begin{tabular}{p{\textwidth}} 
\toprule
\rowcolor[gray]{0.9}
\textit{Prompt for Qwen2.5-14B-Instruct to determine whether an edit instruction pertains to the main subject of an image} \\ 
\midrule
\textbf{Prompt:} You are an assistant that determines whether an edit instruction pertains to the main subject of an image.\\
The main subject refers to humans or animals only. If the edit instruction is related to the main subject, respond with \texttt{'yes'}.\\
If it pertains to background, large areas of vegetation, or environmental information, respond with \texttt{'no'}.\\
Here are some examples:\\
1. replace the grass with sand $\rightarrow$ \texttt{no}\\
2. replace the trees with palm trees $\rightarrow$ \texttt{no}\\
3. replace the dirt road with a cobblestone path $\rightarrow$ \texttt{no}\\
4. replace the bird with a parrot $\rightarrow$ \texttt{yes}\\
Now, analyze the following instruction and respond accordingly: \texttt{\{instruction\}}\\
\bottomrule
\end{tabular}
\label{tab:edit_instruction_main_subject}
\end{table*}

\begin{table*}[h]
\centering
\caption{Prompt designs for extracting and modifying edit instructions based on either replacement or addition of subjects.}
\begin{tabular}{p{\textwidth}} 
\toprule
\rowcolor[gray]{0.9}
\textit{Prompt for extracting the main subject and modifying the edit instruction} \\ 
\midrule
\textbf{Prompt 1: Replace Examples} \\
You are an assistant that determines the main subject of an edit instruction and outputs the extracted main subject along with a modified prompt.\\
Given an edit instruction like ``replace X with Y'':\\
- Extract ``Y'' (the replacement subject).\\
- Output the extracted subject.\\
- Modify the instruction by replacing ``Y'' with \texttt{<imagehere>} in the original prompt and output the new instruction.\\
Examples:\\
1. replace one woman with a man $\rightarrow$ Output: \texttt{a man}, replace one woman with \texttt{<imagehere>}\\
2. replace the castle with another castle $\rightarrow$ Output: \texttt{castle}, replace the castle with \texttt{<imagehere>}\\
3. replace the desk with a white one in the bottom middle $\rightarrow$ Output: \texttt{desk}, replace the desk with \texttt{<imagehere>} in the bottom middle\\
\midrule
\textbf{Prompt 2: Add Examples} \\
You are an assistant that extracts the added subject in the edit instruction.\\
Given an edit instruction like ``add X to the image'':\\
- Extract ``X'' (the added subject).\\
- Output the extracted subject.\\
- Modify the instruction by replacing ``X'' with \texttt{<imagehere>} in the original prompt and output the new instruction.\\
Here are some examples:\\
1. add human over the stone in the bottom left $\rightarrow$ Output: \texttt{human}, add \texttt{<imagehere>} over the stone in the bottom left\\
2. add a car on the road at the bottom $\rightarrow$ Output: \texttt{a car}, add \texttt{<imagehere>} on the road at the bottom\\
3. Add an owl on the left shoulder $\rightarrow$ Output: \texttt{an owl}, Add \texttt{<imagehere>} on the left shoulder\\
\bottomrule
\end{tabular}
\label{tab:edit_instruction_subject_extraction}
\end{table*}

%%%%%%%%%%%%%

\clearpage
\begin{acks}
This work was supported by the National Natural Science Foundation of China (No.62172393), and
Major Public Welfare Project of Henan Province
(No.201300311200).
\end{acks}

\bibliographystyle{ACM-Reference-Format}
\balance
\bibliography{ref}


\begin{thebibliography}{72}

%%% the Web address.

\ifx \showCODEN    \undefined \def \showCODEN     #1{\unskip}     \fi
\ifx \showISBNx    \undefined \def \showISBNx     #1{\unskip}     \fi
\ifx \showISBNxiii \undefined \def \showISBNxiii  #1{\unskip}     \fi
\ifx \showISSN     \undefined \def \showISSN      #1{\unskip}     \fi
\ifx \showLCCN     \undefined \def \showLCCN      #1{\unskip}     \fi
\ifx \shownote     \undefined \def \shownote      #1{#1}          \fi
\ifx \showarticletitle \undefined \def \showarticletitle #1{#1}   \fi
\ifx \showURL      \undefined \def \showURL       {\relax}        \fi
% invisible to TeX
\providecommand\bibfield[2]{#2}
\providecommand\bibinfo[2]{#2}
\providecommand\natexlab[1]{#1}
\providecommand\showeprint[2][]{arXiv:#2}

\bibitem[Abid et~al\mbox{.}(2019)]%
        {abid2019gradio}
\bibfield{author}{\bibinfo{person}{Abubakar Abid}, \bibinfo{person}{Ali
  Abdalla}, \bibinfo{person}{Ali Abid}, \bibinfo{person}{Dawood Khan},
  \bibinfo{person}{Abdulrahman Alfozan}, {and} \bibinfo{person}{James Zou}.}
  \bibinfo{year}{2019}\natexlab{}.
\newblock \showarticletitle{Gradio: Hassle-Free Sharing and Testing of ML
  Models in the Wild}.
\newblock \bibinfo{journal}{\emph{arXiv preprint arXiv:1906.02569}}
  (\bibinfo{year}{2019}).
\newblock


\bibitem[Brooks et~al\mbox{.}(2023)]%
        {brooks2023instructpix2pix}
\bibfield{author}{\bibinfo{person}{Tim Brooks}, \bibinfo{person}{Aleksander
  Holynski}, {and} \bibinfo{person}{Alexei~A Efros}.}
  \bibinfo{year}{2023}\natexlab{}.
\newblock \showarticletitle{Instructpix2pix: Learning to follow image editing
  instructions}. In \bibinfo{booktitle}{\emph{Proceedings of the IEEE/CVF
  Conference on Computer Vision and Pattern Recognition}}.
  \bibinfo{pages}{18392--18402}.
\newblock


\bibitem[Canberk et~al\mbox{.}(2025)]%
        {canberk2025erasedraw}
\bibfield{author}{\bibinfo{person}{Alper Canberk}, \bibinfo{person}{Maksym
  Bondarenko}, \bibinfo{person}{Ege Ozguroglu}, \bibinfo{person}{Ruoshi Liu},
  {and} \bibinfo{person}{Carl Vondrick}.} \bibinfo{year}{2025}\natexlab{}.
\newblock \showarticletitle{EraseDraw: Learning to Insert Objects by Erasing
  Them from Images}. In \bibinfo{booktitle}{\emph{European Conference on
  Computer Vision}}. Springer, \bibinfo{pages}{144--160}.
\newblock


\bibitem[Chen et~al\mbox{.}(2023)]%
        {chen2023pixart}
\bibfield{author}{\bibinfo{person}{Junsong Chen}, \bibinfo{person}{Jincheng
  Yu}, \bibinfo{person}{Chongjian Ge}, \bibinfo{person}{Lewei Yao},
  \bibinfo{person}{Enze Xie}, \bibinfo{person}{Yue Wu},
  \bibinfo{person}{Zhongdao Wang}, \bibinfo{person}{James Kwok},
  \bibinfo{person}{Ping Luo}, \bibinfo{person}{Huchuan Lu}, {et~al\mbox{.}}}
  \bibinfo{year}{2023}\natexlab{}.
\newblock \showarticletitle{Pixart-$\alpha$: Fast training of diffusion
  transformer for photorealistic text-to-image synthesis}.
\newblock \bibinfo{journal}{\emph{arXiv preprint arXiv:2310.00426}}
  (\bibinfo{year}{2023}).
\newblock


\bibitem[Chen et~al\mbox{.}(2024a)]%
        {chen2024zero}
\bibfield{author}{\bibinfo{person}{Xi Chen}, \bibinfo{person}{Yutong Feng},
  \bibinfo{person}{Mengting Chen}, \bibinfo{person}{Yiyang Wang},
  \bibinfo{person}{Shilong Zhang}, \bibinfo{person}{Yu Liu},
  \bibinfo{person}{Yujun Shen}, {and} \bibinfo{person}{Hengshuang Zhao}.}
  \bibinfo{year}{2024}\natexlab{a}.
\newblock \showarticletitle{Zero-shot image editing with reference imitation}.
\newblock \bibinfo{journal}{\emph{Advances in Neural Information Processing
  Systems}}  \bibinfo{volume}{37} (\bibinfo{year}{2024}),
  \bibinfo{pages}{84010--84032}.
\newblock


\bibitem[Chen et~al\mbox{.}(2024b)]%
        {chen2024anydoor}
\bibfield{author}{\bibinfo{person}{Xi Chen}, \bibinfo{person}{Lianghua Huang},
  \bibinfo{person}{Yu Liu}, \bibinfo{person}{Yujun Shen}, \bibinfo{person}{Deli
  Zhao}, {and} \bibinfo{person}{Hengshuang Zhao}.}
  \bibinfo{year}{2024}\natexlab{b}.
\newblock \showarticletitle{Anydoor: Zero-shot object-level image
  customization}. In \bibinfo{booktitle}{\emph{Proceedings of the IEEE/CVF
  conference on computer vision and pattern recognition}}.
  \bibinfo{pages}{6593--6602}.
\newblock


\bibitem[Chen et~al\mbox{.}(2024c)]%
        {chen2024unireal}
\bibfield{author}{\bibinfo{person}{Xi Chen}, \bibinfo{person}{Zhifei Zhang},
  \bibinfo{person}{He Zhang}, \bibinfo{person}{Yuqian Zhou},
  \bibinfo{person}{Soo~Ye Kim}, \bibinfo{person}{Qing Liu},
  \bibinfo{person}{Yijun Li}, \bibinfo{person}{Jianming Zhang},
  \bibinfo{person}{Nanxuan Zhao}, \bibinfo{person}{Yilin Wang},
  {et~al\mbox{.}}} \bibinfo{year}{2024}\natexlab{c}.
\newblock \showarticletitle{UniReal: Universal Image Generation and Editing via
  Learning Real-world Dynamics}.
\newblock \bibinfo{journal}{\emph{arXiv preprint arXiv:2412.07774}}
  (\bibinfo{year}{2024}).
\newblock


\bibitem[Choi et~al\mbox{.}(2024)]%
        {choi2024improving}
\bibfield{author}{\bibinfo{person}{Yisol Choi}, \bibinfo{person}{Sangkyung
  Kwak}, \bibinfo{person}{Kyungmin Lee}, \bibinfo{person}{Hyungwon Choi}, {and}
  \bibinfo{person}{Jinwoo Shin}.} \bibinfo{year}{2024}\natexlab{}.
\newblock \showarticletitle{Improving Diffusion Models for Authentic Virtual
  Try-on in the Wild}.
\newblock \bibinfo{journal}{\emph{arXiv preprint arXiv:2403.05139}}
  (\bibinfo{year}{2024}).
\newblock


\bibitem[Chung et~al\mbox{.}(2024)]%
        {chung2024scaling}
\bibfield{author}{\bibinfo{person}{Hyung~Won Chung}, \bibinfo{person}{Le Hou},
  \bibinfo{person}{Shayne Longpre}, \bibinfo{person}{Barret Zoph},
  \bibinfo{person}{Yi Tay}, \bibinfo{person}{William Fedus},
  \bibinfo{person}{Yunxuan Li}, \bibinfo{person}{Xuezhi Wang},
  \bibinfo{person}{Mostafa Dehghani}, \bibinfo{person}{Siddhartha Brahma},
  {et~al\mbox{.}}} \bibinfo{year}{2024}\natexlab{}.
\newblock \showarticletitle{Scaling instruction-finetuned language models}.
\newblock \bibinfo{journal}{\emph{Journal of Machine Learning Research}}
  \bibinfo{volume}{25}, \bibinfo{number}{70} (\bibinfo{year}{2024}),
  \bibinfo{pages}{1--53}.
\newblock


\bibitem[Esser et~al\mbox{.}(2024)]%
        {esser2024scaling}
\bibfield{author}{\bibinfo{person}{Patrick Esser}, \bibinfo{person}{Sumith
  Kulal}, \bibinfo{person}{Andreas Blattmann}, \bibinfo{person}{Rahim
  Entezari}, \bibinfo{person}{Jonas M{\"u}ller}, \bibinfo{person}{Harry Saini},
  \bibinfo{person}{Yam Levi}, \bibinfo{person}{Dominik Lorenz},
  \bibinfo{person}{Axel Sauer}, \bibinfo{person}{Frederic Boesel},
  {et~al\mbox{.}}} \bibinfo{year}{2024}\natexlab{}.
\newblock \showarticletitle{Scaling rectified flow transformers for
  high-resolution image synthesis}. In \bibinfo{booktitle}{\emph{Forty-first
  International Conference on Machine Learning}}.
\newblock


\bibitem[Fang et~al\mbox{.}(2023)]%
        {fang2023eva}
\bibfield{author}{\bibinfo{person}{Yuxin Fang}, \bibinfo{person}{Wen Wang},
  \bibinfo{person}{Binhui Xie}, \bibinfo{person}{Quan Sun},
  \bibinfo{person}{Ledell Wu}, \bibinfo{person}{Xinggang Wang},
  \bibinfo{person}{Tiejun Huang}, \bibinfo{person}{Xinlong Wang}, {and}
  \bibinfo{person}{Yue Cao}.} \bibinfo{year}{2023}\natexlab{}.
\newblock \showarticletitle{Eva: Exploring the limits of masked visual
  representation learning at scale}. In \bibinfo{booktitle}{\emph{Proceedings
  of the IEEE/CVF Conference on Computer Vision and Pattern Recognition}}.
  \bibinfo{pages}{19358--19369}.
\newblock


\bibitem[Fu et~al\mbox{.}(2023)]%
        {fu2023guiding}
\bibfield{author}{\bibinfo{person}{Tsu-Jui Fu}, \bibinfo{person}{Wenze Hu},
  \bibinfo{person}{Xianzhi Du}, \bibinfo{person}{William~Yang Wang},
  \bibinfo{person}{Yinfei Yang}, {and} \bibinfo{person}{Zhe Gan}.}
  \bibinfo{year}{2023}\natexlab{}.
\newblock \showarticletitle{Guiding instruction-based image editing via
  multimodal large language models}.
\newblock \bibinfo{journal}{\emph{arXiv preprint arXiv:2309.17102}}
  (\bibinfo{year}{2023}).
\newblock


\bibitem[Ge et~al\mbox{.}(2024)]%
        {ge2024seed}
\bibfield{author}{\bibinfo{person}{Yuying Ge}, \bibinfo{person}{Sijie Zhao},
  \bibinfo{person}{Chen Li}, \bibinfo{person}{Yixiao Ge}, {and}
  \bibinfo{person}{Ying Shan}.} \bibinfo{year}{2024}\natexlab{}.
\newblock \showarticletitle{SEED-Data-Edit Technical Report: A Hybrid Dataset
  for Instructional Image Editing}.
\newblock \bibinfo{journal}{\emph{arXiv preprint arXiv:2405.04007}}
  (\bibinfo{year}{2024}).
\newblock


\bibitem[Glorot et~al\mbox{.}(2011)]%
        {glorot2011deep}
\bibfield{author}{\bibinfo{person}{Xavier Glorot}, \bibinfo{person}{Antoine
  Bordes}, {and} \bibinfo{person}{Yoshua Bengio}.}
  \bibinfo{year}{2011}\natexlab{}.
\newblock \showarticletitle{Deep sparse rectifier neural networks}. In
  \bibinfo{booktitle}{\emph{Proceedings of the fourteenth international
  conference on artificial intelligence and statistics}}. JMLR Workshop and
  Conference Proceedings, \bibinfo{pages}{315--323}.
\newblock


\bibitem[Han et~al\mbox{.}(2024)]%
        {han2024ace}
\bibfield{author}{\bibinfo{person}{Zhen Han}, \bibinfo{person}{Zeyinzi Jiang},
  \bibinfo{person}{Yulin Pan}, \bibinfo{person}{Jingfeng Zhang},
  \bibinfo{person}{Chaojie Mao}, \bibinfo{person}{Chenwei Xie},
  \bibinfo{person}{Yu Liu}, {and} \bibinfo{person}{Jingren Zhou}.}
  \bibinfo{year}{2024}\natexlab{}.
\newblock \showarticletitle{ACE: All-round Creator and Editor Following
  Instructions via Diffusion Transformer}.
\newblock \bibinfo{journal}{\emph{arXiv preprint arXiv:2410.00086}}
  (\bibinfo{year}{2024}).
\newblock


\bibitem[He et~al\mbox{.}(2024a)]%
        {he2024affordance}
\bibfield{author}{\bibinfo{person}{Jixuan He}, \bibinfo{person}{Wanhua Li},
  \bibinfo{person}{Ye Liu}, \bibinfo{person}{Junsik Kim},
  \bibinfo{person}{Donglai Wei}, {and} \bibinfo{person}{Hanspeter Pfister}.}
  \bibinfo{year}{2024}\natexlab{a}.
\newblock \showarticletitle{Affordance-Aware Object Insertion via Mask-Aware
  Dual Diffusion}.
\newblock \bibinfo{journal}{\emph{arXiv preprint arXiv:2412.14462}}
  (\bibinfo{year}{2024}).
\newblock


\bibitem[He et~al\mbox{.}(2024b)]%
        {he2024freeedit}
\bibfield{author}{\bibinfo{person}{Runze He}, \bibinfo{person}{Kai Ma},
  \bibinfo{person}{Linjiang Huang}, \bibinfo{person}{Shaofei Huang},
  \bibinfo{person}{Jialin Gao}, \bibinfo{person}{Xiaoming Wei},
  \bibinfo{person}{Jiao Dai}, \bibinfo{person}{Jizhong Han}, {and}
  \bibinfo{person}{Si Liu}.} \bibinfo{year}{2024}\natexlab{b}.
\newblock \showarticletitle{Freeedit: Mask-free reference-based image editing
  with multi-modal instruction}.
\newblock \bibinfo{journal}{\emph{arXiv preprint arXiv:2409.18071}}
  (\bibinfo{year}{2024}).
\newblock


\bibitem[Hsiao et~al\mbox{.}(2025)]%
        {hsiao2025tf}
\bibfield{author}{\bibinfo{person}{Teng-Fang Hsiao}, \bibinfo{person}{Bo-Kai
  Ruan}, \bibinfo{person}{Yi-Lun Wu}, \bibinfo{person}{Tzu-Ling Lin}, {and}
  \bibinfo{person}{Hong-Han Shuai}.} \bibinfo{year}{2025}\natexlab{}.
\newblock \showarticletitle{TF-TI2I: Training-Free Text-and-Image-to-Image
  Generation via Multi-Modal Implicit-Context Learning in Text-to-Image
  Models}.
\newblock \bibinfo{journal}{\emph{arXiv preprint arXiv:2503.15283}}
  (\bibinfo{year}{2025}).
\newblock


\bibitem[Hu et~al\mbox{.}(2024)]%
        {hu2024instruct}
\bibfield{author}{\bibinfo{person}{Hexiang Hu}, \bibinfo{person}{Kelvin~CK
  Chan}, \bibinfo{person}{Yu-Chuan Su}, \bibinfo{person}{Wenhu Chen},
  \bibinfo{person}{Yandong Li}, \bibinfo{person}{Kihyuk Sohn},
  \bibinfo{person}{Yang Zhao}, \bibinfo{person}{Xue Ben},
  \bibinfo{person}{Boqing Gong}, \bibinfo{person}{William Cohen},
  {et~al\mbox{.}}} \bibinfo{year}{2024}\natexlab{}.
\newblock \showarticletitle{Instruct-Imagen: Image generation with multi-modal
  instruction}. In \bibinfo{booktitle}{\emph{Proceedings of the IEEE/CVF
  Conference on Computer Vision and Pattern Recognition}}.
  \bibinfo{pages}{4754--4763}.
\newblock


\bibitem[Huang et~al\mbox{.}(2024a)]%
        {huang2024group}
\bibfield{author}{\bibinfo{person}{Lianghua Huang}, \bibinfo{person}{Wei Wang},
  \bibinfo{person}{Zhi-Fan Wu}, \bibinfo{person}{Huanzhang Dou},
  \bibinfo{person}{Yupeng Shi}, \bibinfo{person}{Yutong Feng},
  \bibinfo{person}{Chen Liang}, \bibinfo{person}{Yu Liu}, {and}
  \bibinfo{person}{Jingren Zhou}.} \bibinfo{year}{2024}\natexlab{a}.
\newblock \showarticletitle{Group diffusion transformers are unsupervised
  multitask learners}.
\newblock  (\bibinfo{year}{2024}).
\newblock


\bibitem[Huang et~al\mbox{.}(2024b)]%
        {huang2024smartedit}
\bibfield{author}{\bibinfo{person}{Yuzhou Huang}, \bibinfo{person}{Liangbin
  Xie}, \bibinfo{person}{Xintao Wang}, \bibinfo{person}{Ziyang Yuan},
  \bibinfo{person}{Xiaodong Cun}, \bibinfo{person}{Yixiao Ge},
  \bibinfo{person}{Jiantao Zhou}, \bibinfo{person}{Chao Dong},
  \bibinfo{person}{Rui Huang}, \bibinfo{person}{Ruimao Zhang}, {et~al\mbox{.}}}
  \bibinfo{year}{2024}\natexlab{b}.
\newblock \showarticletitle{Smartedit: Exploring complex instruction-based
  image editing with multimodal large language models}. In
  \bibinfo{booktitle}{\emph{Proceedings of the IEEE/CVF Conference on Computer
  Vision and Pattern Recognition}}. \bibinfo{pages}{8362--8371}.
\newblock


\bibitem[Huang et~al\mbox{.}(2025)]%
        {huang2025wegen}
\bibfield{author}{\bibinfo{person}{Zhipeng Huang}, \bibinfo{person}{Shaobin
  Zhuang}, \bibinfo{person}{Canmiao Fu}, \bibinfo{person}{Binxin Yang},
  \bibinfo{person}{Ying Zhang}, \bibinfo{person}{Chong Sun},
  \bibinfo{person}{Zhizheng Zhang}, \bibinfo{person}{Yali Wang},
  \bibinfo{person}{Chen Li}, {and} \bibinfo{person}{Zheng-Jun Zha}.}
  \bibinfo{year}{2025}\natexlab{}.
\newblock \showarticletitle{WeGen: A Unified Model for Interactive Multimodal
  Generation as We Chat}.
\newblock \bibinfo{journal}{\emph{arXiv preprint arXiv:2503.01115}}
  (\bibinfo{year}{2025}).
\newblock


\bibitem[Hui et~al\mbox{.}(2024)]%
        {hui2024hq}
\bibfield{author}{\bibinfo{person}{Mude Hui}, \bibinfo{person}{Siwei Yang},
  \bibinfo{person}{Bingchen Zhao}, \bibinfo{person}{Yichun Shi},
  \bibinfo{person}{Heng Wang}, \bibinfo{person}{Peng Wang},
  \bibinfo{person}{Yuyin Zhou}, {and} \bibinfo{person}{Cihang Xie}.}
  \bibinfo{year}{2024}\natexlab{}.
\newblock \showarticletitle{Hq-edit: A high-quality dataset for
  instruction-based image editing}.
\newblock \bibinfo{journal}{\emph{arXiv preprint arXiv:2404.09990}}
  (\bibinfo{year}{2024}).
\newblock


\bibitem[Hurst et~al\mbox{.}(2024)]%
        {hurst2024gpt}
\bibfield{author}{\bibinfo{person}{Aaron Hurst}, \bibinfo{person}{Adam Lerer},
  \bibinfo{person}{Adam~P Goucher}, \bibinfo{person}{Adam Perelman},
  \bibinfo{person}{Aditya Ramesh}, \bibinfo{person}{Aidan Clark},
  \bibinfo{person}{AJ Ostrow}, \bibinfo{person}{Akila Welihinda},
  \bibinfo{person}{Alan Hayes}, \bibinfo{person}{Alec Radford},
  {et~al\mbox{.}}} \bibinfo{year}{2024}\natexlab{}.
\newblock \showarticletitle{Gpt-4o system card}.
\newblock \bibinfo{journal}{\emph{arXiv preprint arXiv:2410.21276}}
  (\bibinfo{year}{2024}).
\newblock


\bibitem[Kang et~al\mbox{.}(2025)]%
        {kang2025flux}
\bibfield{author}{\bibinfo{person}{Hao Kang}, \bibinfo{person}{Stathi
  Fotiadis}, \bibinfo{person}{Liming Jiang}, \bibinfo{person}{Qing Yan},
  \bibinfo{person}{Yumin Jia}, \bibinfo{person}{Zichuan Liu},
  \bibinfo{person}{Min~Jin Chong}, {and} \bibinfo{person}{Xin Lu}.}
  \bibinfo{year}{2025}\natexlab{}.
\newblock \showarticletitle{Flux Already Knows--Activating Subject-Driven Image
  Generation without Training}.
\newblock \bibinfo{journal}{\emph{arXiv preprint arXiv:2504.11478}}
  (\bibinfo{year}{2025}).
\newblock


\bibitem[Kim et~al\mbox{.}(2024)]%
        {kim2024learning}
\bibfield{author}{\bibinfo{person}{Taewook Kim}, \bibinfo{person}{Wei Chen},
  {and} \bibinfo{person}{Qiang Qiu}.} \bibinfo{year}{2024}\natexlab{}.
\newblock \showarticletitle{Learning to Customize Text-to-Image Diffusion In
  Diverse Context}.
\newblock \bibinfo{journal}{\emph{arXiv preprint arXiv:2410.10058}}
  (\bibinfo{year}{2024}).
\newblock


\bibitem[Kingma(2013)]%
        {kingma2013auto}
\bibfield{author}{\bibinfo{person}{Diederik~P Kingma}.}
  \bibinfo{year}{2013}\natexlab{}.
\newblock \showarticletitle{Auto-encoding variational bayes}.
\newblock \bibinfo{journal}{\emph{arXiv preprint arXiv:1312.6114}}
  (\bibinfo{year}{2013}).
\newblock


\bibitem[Kirillov et~al\mbox{.}(2023)]%
        {kirillov2023segment}
\bibfield{author}{\bibinfo{person}{Alexander Kirillov}, \bibinfo{person}{Eric
  Mintun}, \bibinfo{person}{Nikhila Ravi}, \bibinfo{person}{Hanzi Mao},
  \bibinfo{person}{Chloe Rolland}, \bibinfo{person}{Laura Gustafson},
  \bibinfo{person}{Tete Xiao}, \bibinfo{person}{Spencer Whitehead},
  \bibinfo{person}{Alexander~C Berg}, \bibinfo{person}{Wan-Yen Lo},
  {et~al\mbox{.}}} \bibinfo{year}{2023}\natexlab{}.
\newblock \showarticletitle{Segment anything}. In
  \bibinfo{booktitle}{\emph{Proceedings of the IEEE/CVF International
  Conference on Computer Vision}}. \bibinfo{pages}{4015--4026}.
\newblock


\bibitem[Labs(2023)]%
        {flux2023}
\bibfield{author}{\bibinfo{person}{Black~Forest Labs}.}
  \bibinfo{year}{2023}\natexlab{}.
\newblock \bibinfo{title}{FLUX}.
\newblock
  \bibinfo{howpublished}{\url{https://github.com/black-forest-labs/flux}}.
\newblock


\bibitem[Le et~al\mbox{.}(2024)]%
        {le2024one}
\bibfield{author}{\bibinfo{person}{Duong~H Le}, \bibinfo{person}{Tuan Pham},
  \bibinfo{person}{Sangho Lee}, \bibinfo{person}{Christopher Clark},
  \bibinfo{person}{Aniruddha Kembhavi}, \bibinfo{person}{Stephan Mandt},
  \bibinfo{person}{Ranjay Krishna}, {and} \bibinfo{person}{Jiasen Lu}.}
  \bibinfo{year}{2024}\natexlab{}.
\newblock \showarticletitle{One Diffusion to Generate Them All}.
\newblock \bibinfo{journal}{\emph{arXiv preprint arXiv:2411.16318}}
  (\bibinfo{year}{2024}).
\newblock


\bibitem[Li et~al\mbox{.}(2024a)]%
        {li2024blip}
\bibfield{author}{\bibinfo{person}{Dongxu Li}, \bibinfo{person}{Junnan Li},
  {and} \bibinfo{person}{Steven Hoi}.} \bibinfo{year}{2024}\natexlab{a}.
\newblock \showarticletitle{Blip-diffusion: Pre-trained subject representation
  for controllable text-to-image generation and editing}.
\newblock \bibinfo{journal}{\emph{Advances in Neural Information Processing
  Systems}}  \bibinfo{volume}{36} (\bibinfo{year}{2024}).
\newblock


\bibitem[Li et~al\mbox{.}(2023b)]%
        {li2023blip}
\bibfield{author}{\bibinfo{person}{Junnan Li}, \bibinfo{person}{Dongxu Li},
  \bibinfo{person}{Silvio Savarese}, {and} \bibinfo{person}{Steven Hoi}.}
  \bibinfo{year}{2023}\natexlab{b}.
\newblock \showarticletitle{Blip-2: Bootstrapping language-image pre-training
  with frozen image encoders and large language models}. In
  \bibinfo{booktitle}{\emph{International conference on machine learning}}.
  PMLR, \bibinfo{pages}{19730--19742}.
\newblock


\bibitem[Li et~al\mbox{.}(2025c)]%
        {li2025tuning}
\bibfield{author}{\bibinfo{person}{Pengzhi Li}, \bibinfo{person}{Qiang Nie},
  \bibinfo{person}{Ying Chen}, \bibinfo{person}{Xi Jiang}, \bibinfo{person}{Kai
  Wu}, \bibinfo{person}{Yuhuan Lin}, \bibinfo{person}{Yong Liu},
  \bibinfo{person}{Jinlong Peng}, \bibinfo{person}{Chengjie Wang}, {and}
  \bibinfo{person}{Feng Zheng}.} \bibinfo{year}{2025}\natexlab{c}.
\newblock \showarticletitle{Tuning-free image customization with image and text
  guidance}. In \bibinfo{booktitle}{\emph{European Conference on Computer
  Vision}}. Springer, \bibinfo{pages}{233--250}.
\newblock


\bibitem[Li et~al\mbox{.}(2023a)]%
        {li2023dreamedit}
\bibfield{author}{\bibinfo{person}{Tianle Li}, \bibinfo{person}{Max Ku},
  \bibinfo{person}{Cong Wei}, {and} \bibinfo{person}{Wenhu Chen}.}
  \bibinfo{year}{2023}\natexlab{a}.
\newblock \showarticletitle{Dreamedit: Subject-driven image editing}.
\newblock \bibinfo{journal}{\emph{arXiv preprint arXiv:2306.12624}}
  (\bibinfo{year}{2023}).
\newblock


\bibitem[Li et~al\mbox{.}(2024c)]%
        {li-etal-2024-unimo}
\bibfield{author}{\bibinfo{person}{Wei Li}, \bibinfo{person}{Xue Xu},
  \bibinfo{person}{Jiachen Liu}, {and} \bibinfo{person}{Xinyan Xiao}.}
  \bibinfo{year}{2024}\natexlab{c}.
\newblock \showarticletitle{{UNIMO}-{G}: Unified Image Generation through
  Multimodal Conditional Diffusion}. In \bibinfo{booktitle}{\emph{Proceedings
  of the 62nd Annual Meeting of the Association for Computational Linguistics
  (Volume 1: Long Papers)}}, \bibfield{editor}{\bibinfo{person}{Lun-Wei Ku},
  \bibinfo{person}{Andre Martins}, {and} \bibinfo{person}{Vivek Srikumar}}
  (Eds.). \bibinfo{publisher}{Association for Computational Linguistics},
  \bibinfo{address}{Bangkok, Thailand}, \bibinfo{pages}{6173--6188}.
\newblock
\href{https://doi.org/10.18653/v1/2024.acl-long.335}{doi:\nolinkurl{10.18653/v1/2024.acl-long.335}}


\bibitem[Li et~al\mbox{.}(2024d)]%
        {li2024tokenpacker}
\bibfield{author}{\bibinfo{person}{Wentong Li}, \bibinfo{person}{Yuqian Yuan},
  \bibinfo{person}{Jian Liu}, \bibinfo{person}{Dongqi Tang},
  \bibinfo{person}{Song Wang}, \bibinfo{person}{Jie Qin},
  \bibinfo{person}{Jianke Zhu}, {and} \bibinfo{person}{Lei Zhang}.}
  \bibinfo{year}{2024}\natexlab{d}.
\newblock \showarticletitle{Tokenpacker: Efficient visual projector for
  multimodal llm}.
\newblock \bibinfo{journal}{\emph{arXiv preprint arXiv:2407.02392}}
  (\bibinfo{year}{2024}).
\newblock


\bibitem[Li et~al\mbox{.}(2025b)]%
        {li2025blobctrl}
\bibfield{author}{\bibinfo{person}{Yaowei Li}, \bibinfo{person}{Lingen Li},
  \bibinfo{person}{Zhaoyang Zhang}, \bibinfo{person}{Xiaoyu Li},
  \bibinfo{person}{Guangzhi Wang}, \bibinfo{person}{Hongxiang Li},
  \bibinfo{person}{Xiaodong Cun}, \bibinfo{person}{Ying Shan}, {and}
  \bibinfo{person}{Yuexian Zou}.} \bibinfo{year}{2025}\natexlab{b}.
\newblock \showarticletitle{Blobctrl: A unified and flexible framework for
  element-level image generation and editing}.
\newblock \bibinfo{journal}{\emph{arXiv preprint arXiv:2503.13434}}
  (\bibinfo{year}{2025}).
\newblock


\bibitem[Li et~al\mbox{.}(2024b)]%
        {li2024unifiedmllm}
\bibfield{author}{\bibinfo{person}{Zhaowei Li}, \bibinfo{person}{Wei Wang},
  \bibinfo{person}{YiQing Cai}, \bibinfo{person}{Xu Qi},
  \bibinfo{person}{Pengyu Wang}, \bibinfo{person}{Dong Zhang},
  \bibinfo{person}{Hang Song}, \bibinfo{person}{Botian Jiang},
  \bibinfo{person}{Zhida Huang}, {and} \bibinfo{person}{Tao Wang}.}
  \bibinfo{year}{2024}\natexlab{b}.
\newblock \showarticletitle{Unifiedmllm: Enabling unified representation for
  multi-modal multi-tasks with large language model}.
\newblock \bibinfo{journal}{\emph{arXiv preprint arXiv:2408.02503}}
  (\bibinfo{year}{2024}).
\newblock


\bibitem[Li et~al\mbox{.}(2025a)]%
        {li2025visualcloze}
\bibfield{author}{\bibinfo{person}{Zhong-Yu Li}, \bibinfo{person}{Ruoyi Du},
  \bibinfo{person}{Juncheng Yan}, \bibinfo{person}{Le Zhuo},
  \bibinfo{person}{Zhen Li}, \bibinfo{person}{Peng Gao},
  \bibinfo{person}{Zhanyu Ma}, {and} \bibinfo{person}{Ming-Ming Cheng}.}
  \bibinfo{year}{2025}\natexlab{a}.
\newblock \showarticletitle{VisualCloze: A Universal Image Generation Framework
  via Visual In-Context Learning}.
\newblock \bibinfo{journal}{\emph{arXiv preprint arXiv:2504.07960}}
  (\bibinfo{year}{2025}).
\newblock


\bibitem[Liang et~al\mbox{.}(2025)]%
        {liang2025idea}
\bibfield{author}{\bibinfo{person}{Chen Liang}, \bibinfo{person}{Lianghua
  Huang}, \bibinfo{person}{Jingwu Fang}, \bibinfo{person}{Huanzhang Dou},
  \bibinfo{person}{Wei Wang}, \bibinfo{person}{Zhi-Fan Wu},
  \bibinfo{person}{Yupeng Shi}, \bibinfo{person}{Junge Zhang},
  \bibinfo{person}{Xin Zhao}, {and} \bibinfo{person}{Yu Liu}.}
  \bibinfo{year}{2025}\natexlab{}.
\newblock \showarticletitle{IDEA-Bench: How Far are Generative Models from
  Professional Designing?}. In \bibinfo{booktitle}{\emph{Proceedings of the
  Computer Vision and Pattern Recognition Conference}}.
  \bibinfo{pages}{18541--18551}.
\newblock


\bibitem[Lin et~al\mbox{.}(2024)]%
        {lin2024pixwizard}
\bibfield{author}{\bibinfo{person}{Weifeng Lin}, \bibinfo{person}{Xinyu Wei},
  \bibinfo{person}{Renrui Zhang}, \bibinfo{person}{Le Zhuo},
  \bibinfo{person}{Shitian Zhao}, \bibinfo{person}{Siyuan Huang},
  \bibinfo{person}{Junlin Xie}, \bibinfo{person}{Yu Qiao},
  \bibinfo{person}{Peng Gao}, {and} \bibinfo{person}{Hongsheng Li}.}
  \bibinfo{year}{2024}\natexlab{}.
\newblock \showarticletitle{PixWizard: Versatile image-to-image visual
  assistant with open-language instructions}.
\newblock \bibinfo{journal}{\emph{arXiv preprint arXiv:2409.15278}}
  (\bibinfo{year}{2024}).
\newblock


\bibitem[Lin et~al\mbox{.}(2025)]%
        {lin2025realgeneral}
\bibfield{author}{\bibinfo{person}{Yijing Lin}, \bibinfo{person}{Mengqi Huang},
  \bibinfo{person}{Shuhan Zhuang}, {and} \bibinfo{person}{Zhendong Mao}.}
  \bibinfo{year}{2025}\natexlab{}.
\newblock \showarticletitle{RealGeneral: Unifying Visual Generation via
  Temporal In-Context Learning with Video Models}.
\newblock \bibinfo{journal}{\emph{arXiv preprint arXiv:2503.10406}}
  (\bibinfo{year}{2025}).
\newblock


\bibitem[Liu et~al\mbox{.}(2025)]%
        {liu2025grounding}
\bibfield{author}{\bibinfo{person}{Shilong Liu}, \bibinfo{person}{Zhaoyang
  Zeng}, \bibinfo{person}{Tianhe Ren}, \bibinfo{person}{Feng Li},
  \bibinfo{person}{Hao Zhang}, \bibinfo{person}{Jie Yang},
  \bibinfo{person}{Qing Jiang}, \bibinfo{person}{Chunyuan Li},
  \bibinfo{person}{Jianwei Yang}, \bibinfo{person}{Hang Su}, {et~al\mbox{.}}}
  \bibinfo{year}{2025}\natexlab{}.
\newblock \showarticletitle{Grounding dino: Marrying dino with grounded
  pre-training for open-set object detection}. In
  \bibinfo{booktitle}{\emph{European Conference on Computer Vision}}. Springer,
  \bibinfo{pages}{38--55}.
\newblock


\bibitem[Loshchilov(2017)]%
        {loshchilov2017decoupled}
\bibfield{author}{\bibinfo{person}{I Loshchilov}.}
  \bibinfo{year}{2017}\natexlab{}.
\newblock \showarticletitle{Decoupled weight decay regularization}.
\newblock \bibinfo{journal}{\emph{arXiv preprint arXiv:1711.05101}}
  (\bibinfo{year}{2017}).
\newblock


\bibitem[Pan et~al\mbox{.}(2023)]%
        {pan2023kosmos}
\bibfield{author}{\bibinfo{person}{Xichen Pan}, \bibinfo{person}{Li Dong},
  \bibinfo{person}{Shaohan Huang}, \bibinfo{person}{Zhiliang Peng},
  \bibinfo{person}{Wenhu Chen}, {and} \bibinfo{person}{Furu Wei}.}
  \bibinfo{year}{2023}\natexlab{}.
\newblock \showarticletitle{Kosmos-g: Generating images in context with
  multimodal large language models}.
\newblock \bibinfo{journal}{\emph{arXiv preprint arXiv:2310.02992}}
  (\bibinfo{year}{2023}).
\newblock


\bibitem[Peebles and Xie(2023)]%
        {peebles2023scalable}
\bibfield{author}{\bibinfo{person}{William Peebles} {and}
  \bibinfo{person}{Saining Xie}.} \bibinfo{year}{2023}\natexlab{}.
\newblock \showarticletitle{Scalable diffusion models with transformers}. In
  \bibinfo{booktitle}{\emph{Proceedings of the IEEE/CVF International
  Conference on Computer Vision}}. \bibinfo{pages}{4195--4205}.
\newblock


\bibitem[Peng et~al\mbox{.}(2024)]%
        {peng2024dreambench++}
\bibfield{author}{\bibinfo{person}{Yuang Peng}, \bibinfo{person}{Yuxin Cui},
  \bibinfo{person}{Haomiao Tang}, \bibinfo{person}{Zekun Qi},
  \bibinfo{person}{Runpei Dong}, \bibinfo{person}{Jing Bai},
  \bibinfo{person}{Chunrui Han}, \bibinfo{person}{Zheng Ge},
  \bibinfo{person}{Xiangyu Zhang}, {and} \bibinfo{person}{Shu-Tao Xia}.}
  \bibinfo{year}{2024}\natexlab{}.
\newblock \showarticletitle{Dreambench++: A human-aligned benchmark for
  personalized image generation}.
\newblock \bibinfo{journal}{\emph{arXiv preprint arXiv:2406.16855}}
  (\bibinfo{year}{2024}).
\newblock


\bibitem[Radford et~al\mbox{.}(2021)]%
        {radford2021learning}
\bibfield{author}{\bibinfo{person}{Alec Radford}, \bibinfo{person}{Jong~Wook
  Kim}, \bibinfo{person}{Chris Hallacy}, \bibinfo{person}{Aditya Ramesh},
  \bibinfo{person}{Gabriel Goh}, \bibinfo{person}{Sandhini Agarwal},
  \bibinfo{person}{Girish Sastry}, \bibinfo{person}{Amanda Askell},
  \bibinfo{person}{Pamela Mishkin}, \bibinfo{person}{Jack Clark},
  {et~al\mbox{.}}} \bibinfo{year}{2021}\natexlab{}.
\newblock \showarticletitle{Learning transferable visual models from natural
  language supervision}. In \bibinfo{booktitle}{\emph{International conference
  on machine learning}}. PMLR, \bibinfo{pages}{8748--8763}.
\newblock


\bibitem[Ren et~al\mbox{.}(2024)]%
        {ren2024grounded}
\bibfield{author}{\bibinfo{person}{Tianhe Ren}, \bibinfo{person}{Shilong Liu},
  \bibinfo{person}{Ailing Zeng}, \bibinfo{person}{Jing Lin},
  \bibinfo{person}{Kunchang Li}, \bibinfo{person}{He Cao},
  \bibinfo{person}{Jiayu Chen}, \bibinfo{person}{Xinyu Huang},
  \bibinfo{person}{Yukang Chen}, \bibinfo{person}{Feng Yan}, {et~al\mbox{.}}}
  \bibinfo{year}{2024}\natexlab{}.
\newblock \showarticletitle{Grounded sam: Assembling open-world models for
  diverse visual tasks}.
\newblock \bibinfo{journal}{\emph{arXiv preprint arXiv:2401.14159}}
  (\bibinfo{year}{2024}).
\newblock


\bibitem[Rombach et~al\mbox{.}(2022)]%
        {rombach2022high}
\bibfield{author}{\bibinfo{person}{Robin Rombach}, \bibinfo{person}{Andreas
  Blattmann}, \bibinfo{person}{Dominik Lorenz}, \bibinfo{person}{Patrick
  Esser}, {and} \bibinfo{person}{Bj{\"o}rn Ommer}.}
  \bibinfo{year}{2022}\natexlab{}.
\newblock \showarticletitle{High-resolution image synthesis with latent
  diffusion models}. In \bibinfo{booktitle}{\emph{Proceedings of the IEEE/CVF
  conference on computer vision and pattern recognition}}.
  \bibinfo{pages}{10684--10695}.
\newblock


\bibitem[Ruiz et~al\mbox{.}(2023)]%
        {ruiz2023dreambooth}
\bibfield{author}{\bibinfo{person}{Nataniel Ruiz}, \bibinfo{person}{Yuanzhen
  Li}, \bibinfo{person}{Varun Jampani}, \bibinfo{person}{Yael Pritch},
  \bibinfo{person}{Michael Rubinstein}, {and} \bibinfo{person}{Kfir Aberman}.}
  \bibinfo{year}{2023}\natexlab{}.
\newblock \showarticletitle{Dreambooth: Fine tuning text-to-image diffusion
  models for subject-driven generation}. In
  \bibinfo{booktitle}{\emph{Proceedings of the IEEE/CVF conference on computer
  vision and pattern recognition}}. \bibinfo{pages}{22500--22510}.
\newblock


\bibitem[Sheynin et~al\mbox{.}(2024)]%
        {sheynin2024emu}
\bibfield{author}{\bibinfo{person}{Shelly Sheynin}, \bibinfo{person}{Adam
  Polyak}, \bibinfo{person}{Uriel Singer}, \bibinfo{person}{Yuval Kirstain},
  \bibinfo{person}{Amit Zohar}, \bibinfo{person}{Oron Ashual},
  \bibinfo{person}{Devi Parikh}, {and} \bibinfo{person}{Yaniv Taigman}.}
  \bibinfo{year}{2024}\natexlab{}.
\newblock \showarticletitle{Emu edit: Precise image editing via recognition and
  generation tasks}. In \bibinfo{booktitle}{\emph{Proceedings of the IEEE/CVF
  Conference on Computer Vision and Pattern Recognition}}.
  \bibinfo{pages}{8871--8879}.
\newblock


\bibitem[Shi et~al\mbox{.}(2020)]%
        {shi2020benchmark}
\bibfield{author}{\bibinfo{person}{Jing Shi}, \bibinfo{person}{Ning Xu},
  \bibinfo{person}{Trung Bui}, \bibinfo{person}{Franck Dernoncourt},
  \bibinfo{person}{Zheng Wen}, {and} \bibinfo{person}{Chenliang Xu}.}
  \bibinfo{year}{2020}\natexlab{}.
\newblock \showarticletitle{A benchmark and baseline for language-driven image
  editing}. In \bibinfo{booktitle}{\emph{Proceedings of the Asian Conference on
  Computer Vision}}.
\newblock


\bibitem[Sun et~al\mbox{.}(2024)]%
        {sun2024generative}
\bibfield{author}{\bibinfo{person}{Quan Sun}, \bibinfo{person}{Yufeng Cui},
  \bibinfo{person}{Xiaosong Zhang}, \bibinfo{person}{Fan Zhang},
  \bibinfo{person}{Qiying Yu}, \bibinfo{person}{Yueze Wang},
  \bibinfo{person}{Yongming Rao}, \bibinfo{person}{Jingjing Liu},
  \bibinfo{person}{Tiejun Huang}, {and} \bibinfo{person}{Xinlong Wang}.}
  \bibinfo{year}{2024}\natexlab{}.
\newblock \showarticletitle{Generative multimodal models are in-context
  learners}. In \bibinfo{booktitle}{\emph{Proceedings of the IEEE/CVF
  Conference on Computer Vision and Pattern Recognition}}.
  \bibinfo{pages}{14398--14409}.
\newblock


\bibitem[Suvorov et~al\mbox{.}(2022)]%
        {suvorov2022resolution}
\bibfield{author}{\bibinfo{person}{Roman Suvorov}, \bibinfo{person}{Elizaveta
  Logacheva}, \bibinfo{person}{Anton Mashikhin}, \bibinfo{person}{Anastasia
  Remizova}, \bibinfo{person}{Arsenii Ashukha}, \bibinfo{person}{Aleksei
  Silvestrov}, \bibinfo{person}{Naejin Kong}, \bibinfo{person}{Harshith Goka},
  \bibinfo{person}{Kiwoong Park}, {and} \bibinfo{person}{Victor Lempitsky}.}
  \bibinfo{year}{2022}\natexlab{}.
\newblock \showarticletitle{Resolution-robust large mask inpainting with
  fourier convolutions}. In \bibinfo{booktitle}{\emph{Proceedings of the
  IEEE/CVF winter conference on applications of computer vision}}.
  \bibinfo{pages}{2149--2159}.
\newblock


\bibitem[Tan et~al\mbox{.}(2024)]%
        {tan2024ominicontrol}
\bibfield{author}{\bibinfo{person}{Zhenxiong Tan}, \bibinfo{person}{Songhua
  Liu}, \bibinfo{person}{Xingyi Yang}, \bibinfo{person}{Qiaochu Xue}, {and}
  \bibinfo{person}{Xinchao Wang}.} \bibinfo{year}{2024}\natexlab{}.
\newblock \showarticletitle{Ominicontrol: Minimal and universal control for
  diffusion transformer}.
\newblock \bibinfo{journal}{\emph{arXiv preprint arXiv:2411.15098}}
  \bibinfo{volume}{3} (\bibinfo{year}{2024}).
\newblock


\bibitem[Tao et~al\mbox{.}(2024)]%
        {tao2024we}
\bibfield{author}{\bibinfo{person}{Ming Tao}, \bibinfo{person}{Bing-Kun Bao},
  \bibinfo{person}{Yaowei Wang}, {and} \bibinfo{person}{Changsheng Xu}.}
  \bibinfo{year}{2024}\natexlab{}.
\newblock \showarticletitle{Do We Need to Design Specific Diffusion Models for
  Different Tasks? Try ONE-PIC}.
\newblock \bibinfo{journal}{\emph{arXiv preprint arXiv:2412.05619}}
  (\bibinfo{year}{2024}).
\newblock


\bibitem[Team et~al\mbox{.}(2023)]%
        {MosaicML2023Introducing}
\bibfield{author}{\bibinfo{person}{MosaicML~NLP Team} {et~al\mbox{.}}}
  \bibinfo{year}{2023}\natexlab{}.
\newblock \bibinfo{title}{Introducing mpt-7b: A new standard for open-source,
  commercially usable llms}.
\newblock


\bibitem[Team(2024)]%
        {qwen2.5}
\bibfield{author}{\bibinfo{person}{Qwen Team}.}
  \bibinfo{year}{2024}\natexlab{}.
\newblock \bibinfo{title}{Qwen2.5: A Party of Foundation Models}.
\newblock
\urldef\tempurl%
\url{https://qwenlm.github.io/blog/qwen2.5/}
\showURL{%
\tempurl}


\bibitem[Wang et~al\mbox{.}(2024)]%
        {Qwen2-VL}
\bibfield{author}{\bibinfo{person}{Peng Wang}, \bibinfo{person}{Shuai Bai},
  \bibinfo{person}{Sinan Tan}, \bibinfo{person}{Shijie Wang},
  \bibinfo{person}{Zhihao Fan}, \bibinfo{person}{Jinze Bai},
  \bibinfo{person}{Keqin Chen}, \bibinfo{person}{Xuejing Liu},
  \bibinfo{person}{Jialin Wang}, \bibinfo{person}{Wenbin Ge},
  \bibinfo{person}{Yang Fan}, \bibinfo{person}{Kai Dang},
  \bibinfo{person}{Mengfei Du}, \bibinfo{person}{Xuancheng Ren},
  \bibinfo{person}{Rui Men}, \bibinfo{person}{Dayiheng Liu},
  \bibinfo{person}{Chang Zhou}, \bibinfo{person}{Jingren Zhou}, {and}
  \bibinfo{person}{Junyang Lin}.} \bibinfo{year}{2024}\natexlab{}.
\newblock \showarticletitle{Qwen2-VL: Enhancing Vision-Language Model's
  Perception of the World at Any Resolution}.
\newblock \bibinfo{journal}{\emph{arXiv preprint arXiv:2409.12191}}
  (\bibinfo{year}{2024}).
\newblock


\bibitem[Wu et~al\mbox{.}(2025)]%
        {wu2025core}
\bibfield{author}{\bibinfo{person}{Feize Wu}, \bibinfo{person}{Yun Pang},
  \bibinfo{person}{Junyi Zhang}, \bibinfo{person}{Lianyu Pang},
  \bibinfo{person}{Jian Yin}, \bibinfo{person}{Baoquan Zhao},
  \bibinfo{person}{Qing Li}, {and} \bibinfo{person}{Xudong Mao}.}
  \bibinfo{year}{2025}\natexlab{}.
\newblock \showarticletitle{Core: Context-regularized text embedding learning
  for text-to-image personalization}. In \bibinfo{booktitle}{\emph{Proceedings
  of the AAAI Conference on Artificial Intelligence}},
  Vol.~\bibinfo{volume}{39}. \bibinfo{pages}{8377--8385}.
\newblock


\bibitem[Xiao et~al\mbox{.}(2024)]%
        {xiao2024omnigen}
\bibfield{author}{\bibinfo{person}{Shitao Xiao}, \bibinfo{person}{Yueze Wang},
  \bibinfo{person}{Junjie Zhou}, \bibinfo{person}{Huaying Yuan},
  \bibinfo{person}{Xingrun Xing}, \bibinfo{person}{Ruiran Yan},
  \bibinfo{person}{Shuting Wang}, \bibinfo{person}{Tiejun Huang}, {and}
  \bibinfo{person}{Zheng Liu}.} \bibinfo{year}{2024}\natexlab{}.
\newblock \showarticletitle{Omnigen: Unified image generation}.
\newblock \bibinfo{journal}{\emph{arXiv preprint arXiv:2409.11340}}
  (\bibinfo{year}{2024}).
\newblock


\bibitem[Xu et~al\mbox{.}(2024)]%
        {xu2024insightedit}
\bibfield{author}{\bibinfo{person}{Yingjing Xu}, \bibinfo{person}{Jie Kong},
  \bibinfo{person}{Jiazhi Wang}, \bibinfo{person}{Xiao Pan},
  \bibinfo{person}{Bo Lin}, {and} \bibinfo{person}{Qiang Liu}.}
  \bibinfo{year}{2024}\natexlab{}.
\newblock \showarticletitle{InsightEdit: Towards Better Instruction Following
  for Image Editing}.
\newblock \bibinfo{journal}{\emph{arXiv preprint arXiv:2411.17323}}
  (\bibinfo{year}{2024}).
\newblock


\bibitem[Xue et~al\mbox{.}(2024)]%
        {xue2024xgen}
\bibfield{author}{\bibinfo{person}{Le Xue}, \bibinfo{person}{Manli Shu},
  \bibinfo{person}{Anas Awadalla}, \bibinfo{person}{Jun Wang},
  \bibinfo{person}{An Yan}, \bibinfo{person}{Senthil Purushwalkam},
  \bibinfo{person}{Honglu Zhou}, \bibinfo{person}{Viraj Prabhu},
  \bibinfo{person}{Yutong Dai}, \bibinfo{person}{Michael~S Ryoo},
  {et~al\mbox{.}}} \bibinfo{year}{2024}\natexlab{}.
\newblock \showarticletitle{xgen-mm (blip-3): A family of open large multimodal
  models}.
\newblock \bibinfo{journal}{\emph{arXiv preprint arXiv:2408.08872}}
  (\bibinfo{year}{2024}).
\newblock


\bibitem[Yang et~al\mbox{.}(2023)]%
        {yang2023paint}
\bibfield{author}{\bibinfo{person}{Binxin Yang}, \bibinfo{person}{Shuyang Gu},
  \bibinfo{person}{Bo Zhang}, \bibinfo{person}{Ting Zhang},
  \bibinfo{person}{Xuejin Chen}, \bibinfo{person}{Xiaoyan Sun},
  \bibinfo{person}{Dong Chen}, {and} \bibinfo{person}{Fang Wen}.}
  \bibinfo{year}{2023}\natexlab{}.
\newblock \showarticletitle{Paint by example: Exemplar-based image editing with
  diffusion models}. In \bibinfo{booktitle}{\emph{Proceedings of the IEEE/CVF
  Conference on Computer Vision and Pattern Recognition}}.
  \bibinfo{pages}{18381--18391}.
\newblock


\bibitem[Yang et~al\mbox{.}(2024)]%
        {yang2024dreammix}
\bibfield{author}{\bibinfo{person}{Yicheng Yang}, \bibinfo{person}{Pengxiang
  Li}, \bibinfo{person}{Lu Zhang}, \bibinfo{person}{Liqian Ma},
  \bibinfo{person}{Ping Hu}, \bibinfo{person}{Siyu Du}, \bibinfo{person}{Yunzhi
  Zhuge}, \bibinfo{person}{Xu Jia}, {and} \bibinfo{person}{Huchuan Lu}.}
  \bibinfo{year}{2024}\natexlab{}.
\newblock \showarticletitle{DreamMix: Decoupling Object Attributes for Enhanced
  Editability in Customized Image Inpainting}.
\newblock \bibinfo{journal}{\emph{arXiv preprint arXiv:2411.17223}}
  (\bibinfo{year}{2024}).
\newblock


\bibitem[Ye et~al\mbox{.}(2023)]%
        {ye2023ip}
\bibfield{author}{\bibinfo{person}{Hu Ye}, \bibinfo{person}{Jun Zhang},
  \bibinfo{person}{Sibo Liu}, \bibinfo{person}{Xiao Han}, {and}
  \bibinfo{person}{Wei Yang}.} \bibinfo{year}{2023}\natexlab{}.
\newblock \showarticletitle{Ip-adapter: Text compatible image prompt adapter
  for text-to-image diffusion models}.
\newblock \bibinfo{journal}{\emph{arXiv preprint arXiv:2308.06721}}
  (\bibinfo{year}{2023}).
\newblock


\bibitem[Yildirim et~al\mbox{.}(2023)]%
        {yildirim2023inst}
\bibfield{author}{\bibinfo{person}{Ahmet~Burak Yildirim},
  \bibinfo{person}{Vedat Baday}, \bibinfo{person}{Erkut Erdem},
  \bibinfo{person}{Aykut Erdem}, {and} \bibinfo{person}{Aysegul Dundar}.}
  \bibinfo{year}{2023}\natexlab{}.
\newblock \showarticletitle{Inst-inpaint: Instructing to remove objects with
  diffusion models}.
\newblock \bibinfo{journal}{\emph{arXiv preprint arXiv:2304.03246}}
  (\bibinfo{year}{2023}).
\newblock


\bibitem[Yu et~al\mbox{.}(2024)]%
        {yu2024anyedit}
\bibfield{author}{\bibinfo{person}{Qifan Yu}, \bibinfo{person}{Wei Chow},
  \bibinfo{person}{Zhongqi Yue}, \bibinfo{person}{Kaihang Pan},
  \bibinfo{person}{Yang Wu}, \bibinfo{person}{Xiaoyang Wan},
  \bibinfo{person}{Juncheng Li}, \bibinfo{person}{Siliang Tang},
  \bibinfo{person}{Hanwang Zhang}, {and} \bibinfo{person}{Yueting Zhuang}.}
  \bibinfo{year}{2024}\natexlab{}.
\newblock \showarticletitle{AnyEdit: Mastering Unified High-Quality Image
  Editing for Any Idea}.
\newblock \bibinfo{journal}{\emph{arXiv preprint arXiv:2411.15738}}
  (\bibinfo{year}{2024}).
\newblock


\bibitem[Zhang et~al\mbox{.}(2024)]%
        {zhang2024magicbrush}
\bibfield{author}{\bibinfo{person}{Kai Zhang}, \bibinfo{person}{Lingbo Mo},
  \bibinfo{person}{Wenhu Chen}, \bibinfo{person}{Huan Sun}, {and}
  \bibinfo{person}{Yu Su}.} \bibinfo{year}{2024}\natexlab{}.
\newblock \showarticletitle{Magicbrush: A manually annotated dataset for
  instruction-guided image editing}.
\newblock \bibinfo{journal}{\emph{Advances in Neural Information Processing
  Systems}}  \bibinfo{volume}{36} (\bibinfo{year}{2024}).
\newblock


\bibitem[Zhao et~al\mbox{.}(2024)]%
        {zhao2024ultraedit}
\bibfield{author}{\bibinfo{person}{Haozhe Zhao}, \bibinfo{person}{Xiaojian Ma},
  \bibinfo{person}{Liang Chen}, \bibinfo{person}{Shuzheng Si},
  \bibinfo{person}{Rujie Wu}, \bibinfo{person}{Kaikai An},
  \bibinfo{person}{Peiyu Yu}, \bibinfo{person}{Minjia Zhang},
  \bibinfo{person}{Qing Li}, {and} \bibinfo{person}{Baobao Chang}.}
  \bibinfo{year}{2024}\natexlab{}.
\newblock \showarticletitle{UltraEdit: Instruction-based Fine-Grained Image
  Editing at Scale}.
\newblock \bibinfo{journal}{\emph{arXiv preprint arXiv:2407.05282}}
  (\bibinfo{year}{2024}).
\newblock


\bibitem[Zhou et~al\mbox{.}(2025)]%
        {zhou2025fireedit}
\bibfield{author}{\bibinfo{person}{Jun Zhou}, \bibinfo{person}{Jiahao Li},
  \bibinfo{person}{Zunnan Xu}, \bibinfo{person}{Hanhui Li},
  \bibinfo{person}{Yiji Cheng}, \bibinfo{person}{Fa-Ting Hong},
  \bibinfo{person}{Qin Lin}, \bibinfo{person}{Qinglin Lu}, {and}
  \bibinfo{person}{Xiaodan Liang}.} \bibinfo{year}{2025}\natexlab{}.
\newblock \showarticletitle{FireEdit: Fine-grained Instruction-based Image
  Editing via Region-aware Vision Language Model}.
\newblock \bibinfo{journal}{\emph{arXiv preprint arXiv:2503.19839}}
  (\bibinfo{year}{2025}).
\newblock


\end{thebibliography}


\end{document}